\documentclass{article} 
\usepackage{iclr2024_conference,times}

\usepackage{subcaption}
\usepackage{textcomp, gensymb}

\usepackage{amsmath,amsfonts,bm}

\def\eqref#1{equation~\ref{#1}}

\def\1{\bm{1}}

\DeclareMathAlphabet{\mathsfit}{\encodingdefault}{\sfdefault}{m}{sl}
\SetMathAlphabet{\mathsfit}{bold}{\encodingdefault}{\sfdefault}{bx}{n}

\newcommand{\R}{\mathbb{R}}

\def\N{\mathbb{N}}
\def\R{\mathbb{R}}
\def\C{\mathbb{C}}
\def\Z{\mathbb{Z}}
\usepackage{multirow}
\usepackage{booktabs} 
\usepackage{url}
\usepackage{graphicx}
\usepackage{comment}
\usepackage{enumitem}
\def\msets{{\rm msets}}

\usepackage[normalem]{ulem}

\usepackage{hyperref}

\title{\Large Equivariant Matrix Function Neural Networks}

\author{%
Ilyes Batatia$\, ^1$ \ \
Lars L. Schaaf$\, ^1$ \ \
Huajie Chen$\, ^2$ \ \
Gábor Csányi$\, ^1$ \ \
Christoph Ortner$\, ^3$ \ \
Felix A. Faber$\, ^1$  \\
\small{$\, ^1$ University of Cambridge, UK \ \ 
$\, ^2$ Beijing Normal University, China  \ \
$\, ^3$ University of British Columbia, Canada }
}

\iclrfinalcopy 
\begin{document}

\maketitle

\begin{abstract}

Graph Neural Networks (GNNs), especially message-passing neural networks (MPNNs), have emerged as powerful architectures for learning on graphs in diverse applications. However, MPNNs face challenges when modeling non-local interactions in graphs such as large conjugated molecules, and social networks due to oversmoothing and oversquashing. 
Although Spectral GNNs and traditional neural networks such as recurrent neural networks and transformers mitigate these challenges, they often lack generalizability, or fail to capture detailed structural relationships or symmetries in the data. To address these concerns, we introduce Matrix Function Neural Networks (MFNs), a novel architecture that parameterizes \textbf{non-local} interactions through analytic matrix equivariant functions. Employing resolvent expansions offers a straightforward implementation and the potential for linear scaling with system size.
The MFN architecture achieves state-of-the-art performance in standard graph benchmarks, such as the ZINC and TU datasets, and is able to capture intricate non-local interactions in quantum systems, paving the way to new state-of-the-art force fields. The code and the datasets will be made public.

\end{abstract}

\section{Introduction}
\vspace{-3pt}
Graph Neural Networks (GNNs) have proven to be powerful architectures for learning on graphs on a wide range of applications. Various GNN architectures have been proposed, including message passing neural networks (MPNN)~\citep{gilmer2017neural, battaglia2018relational, kipf2016semi, velivckovic2017graph, wu2020comprehensive, deepmind2018protein, hu2019deep, nequip} and higher-order equivariant MPNNs~\citep{Batatia2022mace}. 

MPNNs struggle to model {\bf non-local} interactions both due to computational constraints and over-smoothing~\citep{di2023over}. 
Spectral Graph Neural Networks attempt to address the limitation of this kind by encoding the global structure of a graph using eigenvectors and eigenvalues of a suitable operator. These approaches predominantly focus on Laplacian matrices, exploiting the graph's inherent spectral features. Many spectral GNNs apply polynomial or rational filters~\citep{bianchi2021graph, gasteiger2018predict, wang2022powerful, he2021bernnet, defferrard2016convolutional, zhu2021interpreting, kreuzer2021rethinking} to eigenvalues of graph structures, reaching state-of-the-art accuracy on pure graphs tasks. However, these methods often exhibit rigid architectures that require extensive feature engineering, 
potentially limiting their adaptability to various types of graphs. Moreover, they have been restricted to non-geometric graphs, making them unsuited for molecules and materials.

Traditional neural network architectures, such as recurrent neural networks~\citep{elman1990finding,hochreiter1997long,cho2014properties,graves2013generating} and transformers~\citep{vaswani2017attention} also face challenges when modeling non-local interactions. While transformers can capture some non-local dependencies through their self-attention mechanisms, they come at a significant computational cost due to their quadratic complexity with respect to the input sequence length, even if existing methods can mitigate their cost~\cite{jaegle2022perceiver}. Furthermore, transformers lack inherent structural relationships or positional information within input data, necessitating the use of additional techniques, such as positional encodings~\citep{vaswani2017attention, shaw2018selfattention}. 


In chemistry and material science tasks, some models incorporate classical {\bf long-range} interactions through electrostatics~ \citep{Grisafi2019LODE1, behler2021four_gen_rev, Unke2019PhysNet:Charges}, dispersion, or reciprocal space~\citep{gao2022self_consistent_nn, huguenindumittan2023physicsinspired, kosmala2023ewaldbased}. However, no existing architecture effectively address \textbf{non-local} interactions, where quantum effects can propagate over extensive distances through electronic delocalization, spin coupling, or other many-body non-local mechanisms. This is particularly problematic in systems such as large conjugated molecules, amorphous materials, or metals. 

Consequently, there is a need for new neural network architectures that can efficiently and accurately model complex non-local many-body interactions, while addressing the limitations of current approaches. We propose {\bf Matrix Function Networks} (MFN) as a possible solution to this challenge. Concretely, we make the following contributions.
\vspace{-5pt}
\begin{itemize}
    \item We introduce Matrix Function Networks (MFNs), a new graph neural network architecture able to model non-local interactions in a structured, systematic way.
    \vspace{-2pt}
    \item We introduce the resolvent expansion as a convenient and efficient mechanism to learn a general matrix function. This expansion can in principle be implemented in linear scaling cost with respect to the size of the input.
    \vspace{-2pt}
    \item We demonstrate the ability of our architecture to learn non-local interactions on a dataset of challenging non-local quantum systems, where standard GNNs architectures, including those incorporating global attention, fail to give even qualitatively accurate extrapolative predictions.
    \item We show that MFNs achieve state-of-the-art performance on ZINC and TU graph datasets.
\end{itemize}

\section{Related Work}
\vspace{-5pt}
\paragraph{Overlap matrix fingerprints}~\citep{OMFPs2016} introduced overlap matrix fingerprints (OMFPs), a vector of spectral features of an atomic environment or more generally a point cloud. Given a point cloud, an overlap operator (identity projected on an atomic orbital basis) is constructed, and its ordered eigenvalues (or other invariants) are taken as the features of that point cloud. Although a theoretical understanding of OMFPs is still lacking, computational experiments have shown excellent properties as a distance measure~\citep{OMFPs2016, OMFPs2021}.
\vspace{-10pt}
\paragraph{Spectral Graph Neural Networks}
Spectral GNNs~\citep{WUGNN2021} are GNNs that use spectral filters operating on the Fourier decomposition of the Laplacian operator of the graph. Spectral GNNs are categorized by the type of filters they apply to the spectrum of the Laplacian: ChebyNet~\citep{ChebyConv} approximates the polynomial function of the Laplacian using Chebychev expansions, GPRGNN~\citep{GPRGNN} directly fits coefficients of a fixed polynomial, while ARMA~\citep{ARMA} uses rational filters.
\vspace{-10pt}
\paragraph{Equivariant Neural Networks} Equivariant neural networks are the general class of neural networks that respect certain group symmetries~\citep{Bronstein:2021mdi}. Noteably, convolutional neural networks (CNNs)~\citep{CNNlecun} are equivariant to translations, while $G$-convolutions~\citep{CohenSteerable2016, s.2018spherical, pmlr-v80-kondor18a} generalized CNNs to equivariance of compact groups. Lately, equivariant message passing neural networks~\citep{Anderson2019CormorantCM, Welling2021EGNN, brandstetter2022geometric, nequip, Batatia2022mace, Batatia2022de} have emerged as a powerful architecture for learning on geometric point clouds. Most of these architectures have been shown to lie in a common design space~\cite{Batatia2022de, batatia2023general}.
\vspace{-10pt}
\paragraph{Hamiltonian Learning} A natural application of equivariant neural network architectures is machine learning of (coarse-grained) Hamiltonian operators arising in electronic structure theory. This task of parameterizing the mapping from atomic structures to Hamiltonian operators is currently receiving increasing interest because of the potential extension in accessible observables over purely mechanistic models. The recent works of~\cite{Nigam2022-hamiltonians} and~\cite{2021-acetb1} introduce such parameterizations in terms of a modified equivariant Atomic Cluster Expansion (ACE)~\citep{Drautz2020-tensors}, a precursor to the architecture we employ in the present work. Alternative approaches include~\citep{Hegde2017-ay, Schutt2019-um, Unke-sg-2021, Gu2023-om}.

\vspace{-3pt}
\section{Background}
\vspace{-5pt}
\subsection{Spectral Graph Neural Networks}
\vspace{-5pt}
We briefly review spectral graph neural networks and explain their limitations that our work will overcome.
Consider a graph $\mathcal{G} = (X, \mathcal{E})$ with node set $X$ and edge set $\mathcal{E}$. A graph defined purely by its topology (connectivity) is called a pure graph {\bf}.
Let $n$ denote the number of nodes in the graph, and let $\mathbf{A} \in \R^{n \times n}$ be its adjacency matrix. Define a vector of ones as $\mathbf{1}_{n} \in \R^{n}$. The degree matrix of the graph is $\mathbf{D} = \text{diag}(\mathbf{A} \mathbf{1}_{n})$, and the Laplacian matrix is $\mathbf{L} = \mathbf{D} - \mathbf{A}$.
The Laplacian is a symmetric positive semidefinite matrix and admits a spectral decomposition, $\mathbf{L}  = \mathbf{U} \mathbf{\Lambda} \mathbf{U}^{T}$, where ${\bf U}$ is an orthogonal matrix of eigenvectors and $\mathbf{\Lambda}$ is a diagonal matrix of eigenvalues.

A popular approach to learning functions on graphs is to use convolutional neural networks on the graph.
Spectral graph convolutional networks (SGCNs)  are a class of graph convolutional networks that use the spectral decomposition of the Laplacian matrix to define convolutional filters.
Let $s \in \R^{n}$ be a function of a graph $\mathcal{G}$ and $t$ be a convolutional filter. 
SGCNs take advantage of the spectral decomposition of the Laplacian matrix of graph $\mathcal{G}$ to compute the convolution of $s$ and $t$: 
\begin{equation}
\label{eq:sgnn-convolution}
    s' = t \star_{\mathcal{G}} s = \mathbf{U} ((\mathbf{U}^{T} t) \odot (\mathbf{U}^{T} s))
          = f_{\theta}(\mathbf{L}) s,
\end{equation}
where $f_{\theta}$ is a matrix function of the Laplacian matrix $\mathbf{L}$ of the graph $\mathcal{G}$ and $\odot$ the Hadamard product. 
Various works have proposed different matrix functions $f_{\theta}$, such as polynomial functions~\citep{ChebyConv}, rational functions~\citep{ARMA}, or neural networks~\citep{WUGNN2021}.
Two non-isomorphic graphs can share the same spectrum of their Laplacian operators, while they have different spectra for other graph operators~\citep{JOHNSON198096}.
Therefore, the use of other graph operators as a basis for matrix functions can be beneficial for learning functions on graphs that are strictly more expressive than Laplacian SGCNs.
However, this approach has two main \textbf{limitations}.
\begin{itemize}
    \item \textbf{Expressiveness:} Performing only convolutions with a \textbf{fixed} graph operator, (usually the Laplacian), limits the expressivity of the model. Choosing the most expressive graph operator requires problem-dependent feature engineering.
    \item \textbf{Lie-group symmetries:} The approaches proposed so far have been restricted to \textbf{pure} graphs and in particular do not use additional symmetries for graphs embedded in a vector space. For example, graphs embedded in $\mathbb{R}^3$ often lead to $E(3)$-equivariant learning tasks.
\end{itemize}

To overcome these limitations, we propose MFNs in Section~\ref{sec:mfnns}, 
which allow parameterization of both the graph operator ${\bf H}$ and the matrix function $f_\theta$, and can be formulated to preserve the equivariance under all known group actions. Since a learnable operator ${\bf H}$ prevents the precomputation of diagonalization of ${\bf H}$ during training, we also introduce a method that avoids diagonalization and allows (in principle) linear scaling with the number of nodes.  

\subsection{Equivariant Message Passing Neural Network}
\label{sec:mpnn}
Equivariant Message Passing Neural Networks (MPNNs)~\cite{nequip, Batatia2022mace} are graph neural networks that operate on graphs $\mathcal{G} = (X, \mathcal{E})$ embedded in a vector space $V$. 
The nodes $x_i \in X$ are no longer only a list of indices, but belong to a configuration space $\Omega$ that extends the vector space $V$.
For example, in the atomistic point cloud, $x_{i} = (i, {\bm r}_{i}, \theta_{i}) \in \Omega := \N \times\R^{3} \times \Z$ describing the positions and chemical species of each atom through which the graph is embedded into $V := \mathbb{R}^3$. 
The case of the $\textbf{pure}$ graph can be recovered by setting $\Omega = \N$. 
We are interested in learning graph maps of the form
\begin{equation}
    \Phi \colon \msets(\Omega) \to Z
\end{equation}
where $Z$ is an abstract target space, usually a vector space and $\msets(\Omega)$ the multi-set of states. As the input is a graph, we impose the mapping to be permutation invariant (invariant under relabeling of the nodes). In many applications, the target properties satisfy additional symmetries: When a group $G$ acts on both $\Omega$ (and, therefore, on $\mathcal{G}$) and $Z$, we say that $\Phi$ is $G$-equivariant if,
\begin{equation}
    \Phi \circ g = \rho(g) \Phi \qquad \forall g \in G, 
\end{equation}
where $\rho$ is a representation of the group on the vector space $Z$. A typical strategy is then to embed the nodes $x_i \in X$ into a feature space, where a suitable representation of the group is available.

We represent the state of each node $\sigma_i$ in the layer $t$ of the MPNN by a tuple,
\begin{equation}
  \sigma_{i}^{(t)} = (x_{i}, \boldsymbol{h}_{i}^{(t)}),
\end{equation}
where $x_{i}$ defines the collection of node attributes of the graph as defined previously and $\boldsymbol{h}_{i}^{(t)}$ are its learnable features.
A forward pass of the network consists of multiple \textit{message construction}, \emph{update}, and \emph{readout} steps.
During message construction, a message $\boldsymbol{m}_i^{(t)}$ is created for each node by pooling over its neighbors,
\begin{equation}
  \label{eqn:mpnn-equations}
  \boldsymbol{m}_i^{(t)} = \bigoplus_{j \in \mathcal{N}(i)} M_t\big(\sigma_i^{(t)}, \sigma_j^{(t)}\big), 
  \quad   \boldsymbol{h}_i^{(t+1)} = U_t\big(\sigma_i^{(t)}, \boldsymbol{m}_i^{(t)}\big), 
  \quad   \Phi(\mathcal{G}) = \phi_{\text{out}}\bigg( \Big\{  \big\{ \mathcal{R}_t\big(\sigma^{(t)}_i\big) \big\}_i \Big\}_{t}\bigg).
\end{equation}
where the individual operations have the following meaning: \\[-7mm]
\begin{itemize}
    \item $M_t$ is a learnable message function and
$\bigoplus_{j \in \mathcal{N}(i)}$ is a learnable, permutation invariant pooling operation over the neighbors of atom $i$ (e.g., a sum); \\[-5mm]
    \item $U_t$ is a learnable update function, transforming the message $\boldsymbol{m}_i^{(t)}$ into new features $\boldsymbol{h}_i^{(t+1)}$; \\[-5mm]
    \item $\mathcal{R}_t$ is a learnable node readout, mapping node states $\sigma_i^{(t)}$ to the per-node outputs; \\[-5mm]
    \item $\phi_{\rm out}$ is a global readout map, typically 
    $\phi_{\rm out}(\{\{ \mathcal{R}_{i}^{(t)}\}_i \}_t)
    = \sum_{i, t} \mathcal{R}_i^{(t)}$.
\end{itemize}

%
%
%

Equivariant MPNNs are widely used for learning properties of 3D point clouds, such as molecules and materials. However, there are several limitations to their expressiveness,
\vspace{-8pt}
\begin{itemize}
    \item \textbf{Non-local :} MPNNs are restricted to a finite range, dictated by the number of iterations $T$.
    A large number of properties in physics are long-range, which requires $T \to \infty$. Adding a large number of layers leads to high computational overhead and poor expressivity due to oversmoothness \citep{di2023over}.
    \item \textbf{Correlation order:} Most MPNNs use two body messages, which means that the pooling in Equation~\ref{eqn:mpnn-equations} is just a sum. The MACE architecture~\citep{Batatia2022mace} extended the message construction to an arbitrary order, but increasing the order is still computationally demanding, especially for large $T$.
\end{itemize}
\vspace{-5pt}

\section{Matrix Function Neural Networks}
\vspace{-5pt}
\label{sec:mfnns}
\subsection{The Design Space of Matrix Function Neural Networks}
\label{sec:design-space}
\vspace{-5pt}
\begin{figure}[tp]
    \centering
    \includegraphics[width=1\linewidth]{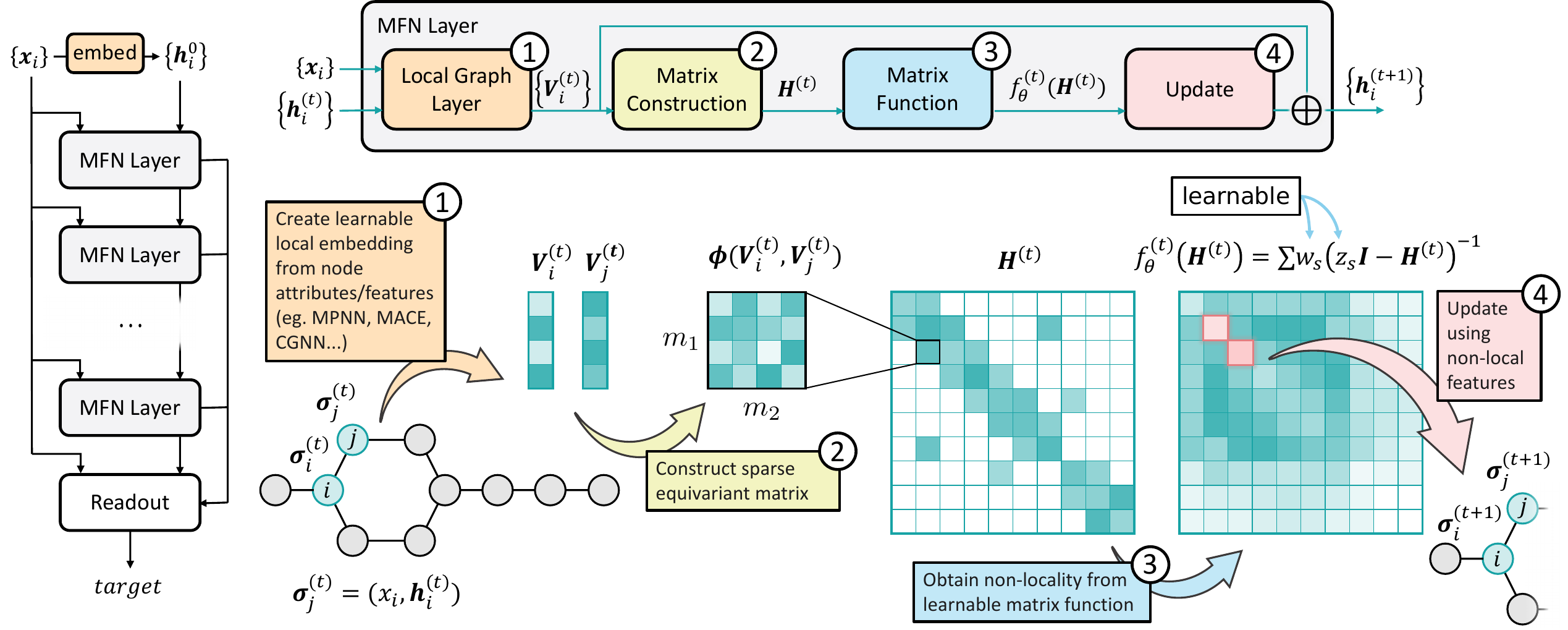}
    \label{fig:enter-label}
	\caption{\textbf{Matrix function network architecture.} Illustrating matrix construction and non-locality of matrix functions on a molecular graph.}
    \label{fig:MFN-framework}
    \vspace{-16pt}
\end{figure}

Our MFN models act in the space of group equivariant matrix operators of the graph, that we denote $\mathcal{H}(\mathcal{G})^{G}$, 
where most commonly $G = O(3)$ the rotations group (including in our examples), but our framework applies to the more general class of reductive Lie groups (e.g., the Lorentz group in high energy physics or  $SU(3)$ in quantum chromodynamics). 
In the same setting as in section~\ref{sec:mpnn}, let $\mathcal{G}$ be an undirected graph with $n$ nodes, and let the states $\sigma_{i}$ be embedded in a configuration space $\Omega$. The architecture operates in four stages at each layer, a \textbf{local graph layer} that learns node features from the states, the \textbf{matrix construction}, the \textbf{matrix function}, and the \textbf{update}. These are visually outlined in figure~\ref{fig:MFN-framework}.

\vspace{-6pt}
\paragraph{Local graph layer} For each iteration $t$, the first step in an MFN layer is to form equivariant node features using a local graph layer (abstractly noted $\mathcal{L}$) as a function of the \textbf{local} environment of a node $i$,
\begin{equation}
\label{eq:local-layer}
    \mathcal{L}^{(t)} : \{\sigma^{(t)}_{j}\}_{j \in \mathcal{N}(i)} \mapsto V^{(t)}_{icm},
\end{equation}
where $V^{(t)}_{icm}$ are the learned equivariant node features and $m$ indexes the representation of $G$ acting on $\textbf{V}$, i.e. $\textbf{V}_{m} \circ g = \sum_{mm'}\rho_{mm'}(g) \textbf{V}_{m'}$. We denote the dimension of the representation by $M$ and the channel dimension by $c$. In the case where $G=O(3)$, the $m$ indices correspond to the usual angular momentum. The specific choice of the graph layer $\mathcal{L}$ depends on the application.
One could use any equivariant architecture to learn these node features, such as a layer of equivariant MPNNs.
In the case of the rotation group, any such function can be approximated with arbitrary accuracy using an ACE or MACE layer~\citep{2021-acetb1, Batatia2022mace}.
This result has been generalized to any reductive Lie group, using the $G$-MACE architecture~\citep{batatia2023general}. It is important to note that the use of more expressive approximators will result in better coverage of the operator space $\mathcal{H}(\mathcal{G})^G$ and therefore better general expressivity.

\paragraph{Matrix construction}
The second step involves constructing a set of graph matrix operators from the learned node features. The space of graph matrix operators, $\mathcal{H}(\mathcal{G})^{G}$, corresponds to the space of operators that are (1) \textbf{self-adjoint}, (2) \textbf{permutation equivariant}, and (3) \textbf{G-equivariant}. The entries of the matrix operator correspond to learnable equivariant edge features expanded in the product of two representations $m_{1}$ and $m_{2}$.  Hence, for each channel $c$, the equivariant operators are square matrices ${\bf H}_{c} \in \C^{Mn \times Mn}$, consistent with $n \times n$ blocks of size  $M\times M$ , where $n$ are the number of nodes. Each nonzero block ${\bf H}_{cij}$ corresponds to learnable edge features between the nodes $i$ and $j$, such that 
\begin{equation}
\label{eq:matrix-construction}
    H_{cij,m_1m_2}^{(t)} = \phi_{cm_1m_2}^{(t)}(\textbf{V}^{(t)}_{i}, \textbf{V}^{(t)}_{j}), \quad \text{if } j \in \mathcal{N}(i) \text{ and } i \in \mathcal{N}(j),
\end{equation}
where $i,j$ denote the indices of the matrix blocks, while $m_1, m_2$ represent the indices within each block. Additionally, $\phi_{cm_1m_2}^{(t)}$ is a learnable equivariant function mapping a tuple of node features to an edge feature. 
For a concrete example of matrix construction in the case of isometry group $O(3)$, see section~\ref{sec:example-mfn} and see Figure~\ref{fig:matrix-o3} for an illustration of the block structure.

The full matrix inherits the equivariance of node features, 
\begin{equation}
\label{eq:matrix-equivariance}
   {\bf H} \circ g = \bm{\rho}(g) {\bf H} \bm{\rho}^{*}(g), \quad \forall g \in G, 
\end{equation}
where $\bm{\rho}$ is a block diagonal unitary matrix that denotes the group action of $G$ on the basis element $\phi$. In the case where $G$ is the rotation group, the blocks of $\bm{\rho}$ correspond to the usual $D$-Wigner matrices.

\paragraph{Matrix function}
The central operation of the MFN architecture is the matrix function, which introduces long-range many-body effects.
Any continuous function $f_\theta : \R \to \R$, with parameters $\theta$, can be used to define a matrix function that maps a square matrix to another square matrix of equal size. Formally, a matrix function on self-adjoint matrices ${\bf H}$ can be defined by its spectral decomposition. Using the fact that ${\bf H}$ is diagonalizable such that ${\bf H} = {\bf U} \bf{\Lambda} {\bf U}^{T}$ with ${\bf U}$ orthogonal and ${\bf \Lambda}$ diagonal,
\begin{equation}
\label{eq:mfn-function}
    f_{\theta}({\bf H}) 
    = {\bf U} f_{\theta}(\bf{\Lambda}) {\bf U}^{T},
\end{equation}
where $f(\bf{\Lambda})$ is a diagonal matrix obtained by applying $f$ to the each diagonal element of $\Lambda$. 
An essential observation is that \textbf{any} continuous matrix function $f_\theta$ preserves equivariance (see proof in Appendix~\ref{sec:equivariant-op}), 
\begin{equation}
\label{eq:mfn-equivariance}
    f_{\theta}({\bf H} \circ g )  = \bm{\rho}(g) f_{\theta}({\bf H}) \bm{\rho^{*}}(g), \quad \forall g \in G.
\end{equation}
%
The matrix function can be related to a high-order many-body equivariant function via the Cayley--Hamilton theorem.
The eigendecomposition in Equation~\ref{eq:mfn-function} is responsible for the non-locality of our approach. 
In practice, computing matrix functions is expensive as it requires diagonalization, scaling as $n^{3}$ with the number of nodes. 
Many approaches are available to approximate matrix functions, such as Chebyshev polynomials or rational approximation, which can leverage potentially cheaper evaluation schemes.
Furthermore, the matrix ${\bf H}$ is sparse in many applications, which can be further exploited to reduce computational cost. 
To further optimize this, we propose to employ a resolvent expansion to parameterize $f_\theta$, detailed in Section~\ref{sec:resolvent}. Similar approaches have been successfully applied to large matrices in other fields such as electronic structure calculations~\citep{pexsi_CMS2009,pexsi_JCPM2013}.

\paragraph{Update}
The \textbf{diagonal update} updates the state of each node with non-local features extracted from the diagonal blocks of the matrix function, 
\begin{equation}
\label{eq:update}
    h^{(t+1)}_{icm} = V^{(t)}_{icm} + \sum_{\tilde{c}} w_{c\tilde{c}}^{(t)} f_{\theta}^{(t)}({\bf H}^{(t)}_{\tilde{c}})_{ii,m0}
\end{equation}
This method is the most computationally efficient since selected inversion techniques~\citep{pexsi_CMS2009} can be employed to efficiently evaluate the diagonal blocks of a matrix function; cf.~Section~\ref{sec:resolvent}. Note that the diagonal blocks are symmetric and therefore extracting $m0$ or $0m$ is equivalent. 
Alternative updates can be defined from the matrix function that we detail in the appendix~\ref{sec:updates-appendix}. The optimal kind of update is a trade-off between expressivity and computational cost. All of the updates differ fundamentally from the standard spectral GNN update, which is a filtering operation. 

The node states are then updated using these new nonlocal node features $\sigma_{i}^{(t+1)} = (x_{i}, \bm{h}^{(t+1)}_{i})$ to form the next states. The steps are repeated for $T$ iterations, starting from the local graph layer.

\paragraph{Readout}
The readout phase is the same as the usual MPNN readout in Equation~\ref{eqn:mpnn-equations}.

\subsection{Resolvent parameterization of matrix functions}
\label{sec:resolvent}
The evaluation of the matrix function in Equation~\ref{eq:mfn-function} is the practical bottleneck of our method.
The cost of the evaluation depends on the choice of parameterization of the univariate function $f_\theta$. 
For a general analytic $\tilde{f} : \mathbb{C} \to \mathbb{C}$, resolvent calculus allows us to represent 
\begin{equation}
    \tilde{f}({\bf H}) = \oint_{\mathcal{C}} \tilde{f}\small(z\small)(z\mathbf{I} - {\bf H})^{-1} \frac{dz}{2\pi i}, 
\end{equation}
where $\mathcal{C}$ is a curve encircling the eigenvalues of ${\bf H}$ and excluding any poles of $\tilde{f}$. Approximating the contour integration with a quadrature rule with nodes $z_s$ and weights $\tilde{w}_s$ yields 
%
    $\tilde{f}({\bf H}) \approx \sum_{s} \tilde{w}_{s}\tilde{f}\small(z_{s}\small)(z_{s}\mathbf{I} - {\bf H})^{-1}$, 
%
and merging $w_s := \tilde{w}_{s}\tilde{f}(z_{s})$ we arrive at the parameterization 
\begin{equation} \label{sec:poleexpansion}
    f_\theta({\bf H}) = \sum_{s} w_{s}(z_{s}\mathbf{I} - {\bf H})^{-1}.
\end{equation}
Pole expansions for evaluating matrix functions have a long and successful history, especially when the arguments are sparse matrices~\citep{higham2008functions}.
The novelty in \eqref{sec:poleexpansion} over standard usage of pole expansions in computing matrix functions~\citep{higham2008functions} is that both the {\em weights} $w_s$ and the {\em poles} $z_s$ are now learnable parameters.

The derivation shows that in the limit of infinitely many pole-weight pairs $(z_s, w_s)$ any analytic matrix function can be represented. Since analytic functions are dense in the space of continuous functions, this means that all continuous matrix functions can be represented in that limit as well (at the same time letting the poles approach the spectrum).
In practice, the poles should be chosen with non-zero \text{Imag}inary parts in order to avoid the spectrum of ${\bf H}$, which is real since ${\bf H}$ is assumed to be self-adjoint. Therefore, we choose adjoint pole weight pairs $(w_s, z_s)$ and $(w_s^*, z_s^*)$ to ensure that $f_\theta$ is real when restricted to real arguments.

\paragraph{Linear scaling cost}
The pole expansion framework is the first key ingredient in linear scaling electronic structure methods~\cite{RevModPhys.71.1085} such as PEXSI~\cite{pexsi_CMS2009,pexsi_JCPM2013}.
The second ingredient is the selected inversion of the resolvent. 
Instead of computing the full inverse, $(z {\bf I} - {\bf H})^{-1}$, one first computes a sparse $LDL^*$ factorization and then selectively computes only the diagonal entries of the resolvent. The bottleneck in this approach is the $LDL^*$ factorization. For a full matrix, it scales as $O(n^3)$ operations and $O(n^2)$ memory. 
The complexity improves considerably for sparse matrices. Suppose that the sparsity pattern is $d$-dimensional, corresponding to a topologically $d$-dimensional graph; e.g. the cumulenes in Section~\ref{sec:cumulenes} are topologically one-dimensional despite being embedded in $\mathbb{R}^3$. Using nested dissection ordering to minimize the fill-in, the cost of the $LDL^*$ factorization reduces to $O(n)$ for $d = 1$ (e.g., natural language processing and quasi-1D molecular systems such as carbon-nano-tubes); $O(n^{3/2})$ operations and $O(n \log n)$ memory for $d = 2$ (e.g., \text{Imag}e recognition); and $O(n^2)$ operations and $O(n^{4/3})$ memory for $d = 3$.  A more formal exposition of this material is given in~\cite{davis2006}. The final step to reduce the computational cost of our architecture to {\em linear scaling} is to replace the $LDL^*$ factorization with an incomplete factorization, as proposed by~\citet{etter2020incomplete} in the context of electronic structure calculations. This would lead to approximations in non-locality that need to be investigated in future work.


\subsection{Example of MFN for the $O(3)$-group}
\label{sec:example-mfn}
\vspace{-5pt}
Here we detail a concrete example for the MFN layer in the case of $O(3)$-group. For concreteness, we will consider point clouds of atoms as input. We will reuse this architecture in the Experiment section.  At each layer $t$, the states of each node are described by a tuple $\sigma_{i} = ({\bf r}_{i}, \theta_{i}, h_{i}^{(t)})$, where $\textbf{r}_{i} \in \R^{3}$ denotes the position of atoms in the 3D space, $\theta_{i} \in \Z$ is the nuclear charge and $h_{i}^{(t)}$ the learnable node features that the MFN layer will update that we initialize just as a learnable embedding of the species. The node features are expanded on the spherical basis and use the usual convention of the $\tilde{l}\tilde{m}$ index (with the correspondence with the single $m$ in Section~\ref{sec:design-space} being $\big (m=0 \to (\tilde{l}=0, \tilde{m}=0) \text{ and } m=1 \to  (\tilde{l}=1, \tilde{m}=-1)$ etc.\big). For the \textbf{local graph layer}, we use a MACE layer (see~\cite{Batatia2022mace} for details) that constructs the local node features $V_{i,c\tilde{l}\tilde{m}}$ in Equation~\ref{eq:local-layer}.
We then construct the matrix using a tensor product of the features of the nodes on nodes $i$ and $j$,
\begin{equation}
    H_{cij, \tilde{l}_{1}\tilde{m}_{1}\tilde{l}_{2}\tilde{m}_{2}} = R_{c}(r_{ij})(V_{i, c\tilde{l}_{1}\tilde{m}_{1}} \otimes V_{j, c\tilde{l}_{2}\tilde{m}_{2}}), \quad \text{if } j \in \mathcal{N}(i) \text{ and } i \in \mathcal{N}(j),
\end{equation}
where $R$ corresponds to a function of the distance $r_{ij}$ between $i$ and $j$. As the graph is undirected, the matrix is symmetric with respect to $i,j$. 
\begin{figure}[h!]
    \centering
\includegraphics[width=0.45\textwidth]{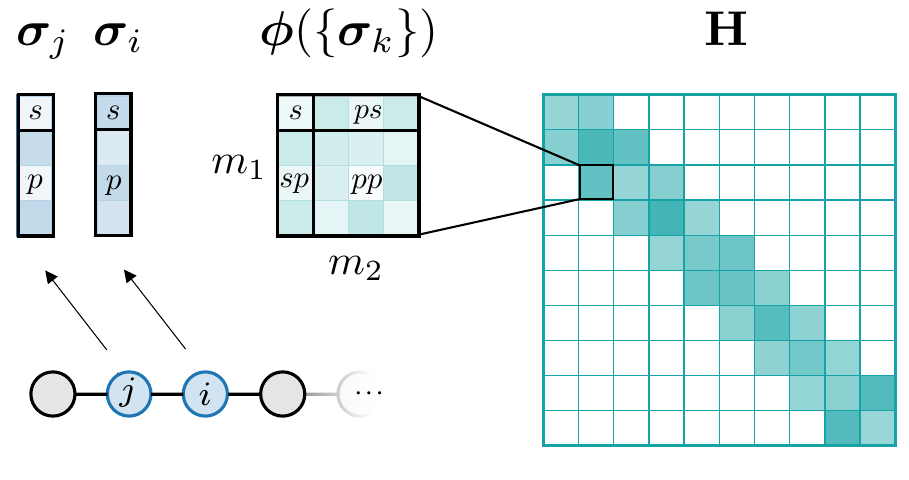}
    \caption{\textbf{Block structure} of a Euclidean MFN operator, ${\bf H}$. Each entry in ${\bf H}$ corresponds to a different product of representations of the group of rotations  $(ss, sp, pp,...)$. Example for L=1.}
    \label{fig:matrix-o3}
\end{figure}
The matrix is sparse, as only the elements corresponding to an edge of the graph are nonzero. This matrix exhibits a block structure illustrated in Fig.~\ref{fig:matrix-o3} similar to Hamiltonian matrices in quantum mechanics with $ss$ and $sp$ orbitials. We refer to $L$ as the maximal value of each spherical index $\tilde{l}_{1}$ and $\tilde{l}_{2}$ of the matrix ($0 \leq \tilde{l}_{1, 2} \leq L$), and we refer to this architecture as $\text{MFN}^{\text{(MACE)}}_{L}$ in the Results section.
\vspace{-8pt}
\subsection{Expressivity of Matrix Function Networks}
%

%
\vspace{-5pt}
Purely formally, one may think of a matrix function $\tilde{f}({\bf H})$ as an infinite power series.
This suggests that MFNs are inherently non-local and exhibit a convolution-like structure, similar to message-passing methods with infinite layers and linear update (see Appendix~\ref{sec:geom-expressiveness} for details). 
This is of interest for modeling causal relationships or non-local interactions by proxy, such as in chemistry or natural language processing. In these cases, the propagation of local effects over long distances results in multiscale effects that are effectively captured by our method. The degree of non-locality of the interaction in MFN can be precisely quantified using the Combes-Thomas theorem~\citep{Combes1973}. We provide a detail analysis in the Appendix~\ref{sec:combes_thomas}. 



\vspace{-6pt}
\section{Results}
\label{sec:resuts}
\vspace{-6pt}
\subsection{Cumulenes: non-local 3D graphs}
\label{sec:cumulenes}
\vspace{-6pt}
We compare the non-locality of MFNs to local MPNNs and global attention MPNNs using linear carbon chains, called cumulenes.  This example is a notorious challenge for state-of-the-art architectures machine learning force fields~\cite{unke2021machine}, and our main motivation for introducing the MFN architecture.

\begin{figure}[h!]
    \centering
    \vspace{-7pt}
    \includegraphics[width=0.99\linewidth]{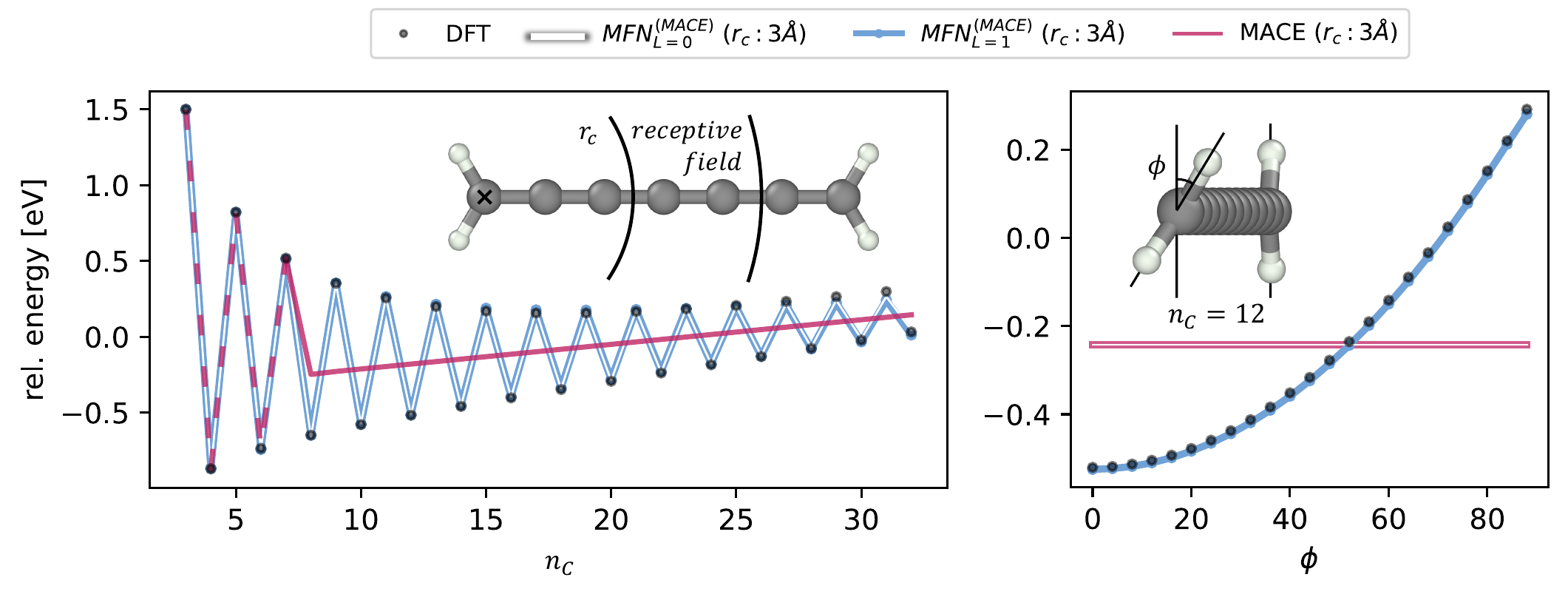}
	\caption{\textbf{Visualizing MFN expressivity on cumulene chains.} The left panel depicts energy trends with respect to cumulene chain length at a fixed angle $\phi=5\degree$. The right panel shows the DFT (ground truth) and the predicted energy as a function of the dihedral angle $\phi$ between the hydrogen atoms for a cumulene chain containing 12 carbon atoms. Local many-body equivariant models (MACE) are only able to capture average trends, even though test configurations are included in the training set. Invariant MFNs ($L=0$) capture only the trends with respect to length, while equivariant MFNs ($L=1$) capture both non-local trends. All models have a cutoff distance $r_c$ of 3\AA{}, corresponding to the nearest neighbors, with two message-passing layers. The cutoff distance as well as MACE's receptive field for the first carbon atom is annotated in the left panel.}
    \label{fig:cumulenes-combined}
\end{figure}

Cumulenes are made up of double-bonded carbon atoms terminated with two hydrogen atoms at each end. Cumulenes exhibit pronounced non-local behavior as a result of strong electron delocalization.  Small changes in chain length and relative angle between the terminating hydrogen atoms can result in large changes in the energy of the system, as visually represented in Figure~\ref{fig:cumulenes-combined}. These structures are known to illustrate the limited expressivity of local models~\citep{unke2021machine} and are similar to the k-chains introduced by~\cite{joshi2023expressive} in the context of the geometric WL test. \cite{frank2022so3krates} showed that global attention is capable of capturing the angular trends of cumulenes with fixed length. We go beyond and demonstrate that MFNs are capable of accurately extrapolating to longer chains, simultaneously capturing length and angle trends. In contrast, global attention models such as Spookynet~\citep{Unke2021}, are unable to extrapolate to longer chains, highlighting the benefit of the matrix function formalism.  For all MFN models in this section, we use MACE layers~\citep{Batatia2022mace} to form the matrix at each layer. We refer to the model as MFN (MACE). Details of the training set and the specific choice of parameterization of the matrix entries are included in the Appendix~\ref{sec:appendix-numerical}.

%
\textbf{Trends with chain length and rotation}
The lowest energy structure of cumulenes alternates between 90- and 0-degree angles for odd and even carbon atom counts, respectively. Consequently, varying the number of carbon atoms at a fixed angle unveils a distinctive zigzag pattern in the ground truth energy (Fig.~\ref{fig:cumulenes-combined} left panel). Although the local model, with two message passing layers, is trained on exactly these configurations, this system is beyond its expressivity, as can be seen by the linear trend for $n_c>7$ in (Figure~\ref{fig:cumulenes-combined} left panel). In contrast, the invariant and equivariant MFN models perfectly reproduce density functional theory (DFT) energies, thanks to their inherent non-locality. 
To demonstrate the necessity of equivariance, we train models on the energy of a fixed size cumulene as a function of the dihedral angle between the hydrogen atoms. Figure~\ref{fig:cumulenes-combined} demonstrates that only the equivariant MFN (L=1) captures the angular trend. 


\textbf{Guaranteed Non-Local dataset}
Datasets designed to test non-local effects often yield unexpectedly high accuracy when evaluated with local models~\citep{kovacsEvaluationMACEForce2023}, complicating the assessment of model non-locality. The dataset introduced here is based on cumulenes, whose strong electronic delocalization results in a directly observable non-locality. The training set contains geometry-optimized cumulenes with 3-10 and 13, 14 carbon atoms, which are then rattled and rotated at various angles. The test set contains cumulenes created in a similar fashion with the same number of carbons (in-domain) and cumulenes of unseen length, not present in the dataset (out-domain 11,12 and 15,16).  Table~\ref{tab:error_table} shows that the MFN architecture significantly outperforms both local and attention-based models (Spookynet). Attention captures some non-locality, resulting in marginally lower errors on the train and in-domain test set. However, the learned non-locality does not generalize to larger molecules, obtaining energy and forces worse than those obtained with a simple local model. The structured non-locality of MFNs enables generalization to larger unseen system sizes.
\begin{table}[htp]
    \centering
    \resizebox{0.79\linewidth}{!}{
    \begin{tabular}{lccccccc}
        \toprule
        \multirow{2}{*}{\textbf{Dataset}} & \multirow{2}{*}{\textbf{$\boldsymbol{n_C}$}}  & \multicolumn{3}{c}{\textbf{E (meV/atom)}} & \multicolumn{3}{c}{\textbf{F (meV/A)}} \\
        & & MACE & \small{SpookyNet}& MFN$^{(\text{MACE})}$ & MACE & \small{SpookyNet} & MFN$^{(\text{MACE})}$ \\
        & & (Local) & (Global attention) & (ours) & (Local) & (Global attention) & (ours) \\
        \midrule
        Train              & 3-10,13,14 & {41.1} &  \underline{31.4} & \textbf{2.0} & 179.6  & \underline{114.1} & \textbf{31.7} \\
        Test (In Domain)   & 3-10,13,14 & {41.8} & \underline{30.8} & \textbf{2.6} & {205.6} & \underline{162.3} &\textbf{34.0} \\
        Test (Out Domain)  & 11,12      & \underline{16.5} & 31.4 & \textbf{0.9} & \underline{108.5} &  {116.2} &\textbf{22.5} \\
        Test (Out Domain)  & 15,16      & \underline{12.0} & {23.4} &\textbf{2.6} & \underline{77.1} & {87.6} &\textbf{37.7} \\
        \bottomrule
    \end{tabular}
    }
    \caption{\textbf{Guaranteed non-local cumulene dataset} containing rattled cumulene chains, with various chain lengths ($n_C$) and hydrogen dihedral angles ($\phi$). The table compares energy (E) and forces (F) RMSEs between local two-layer MPNNs (MACE), global attention MPNNs (SpookyNet), and equivariant MFNs. Train and in-domain test sets contain cumulenes of lengths 3-10 and 13,14. Two out-domain test sets compare different levels of extrapolation to unseen cumulene lengths, containing cumulenes with 11, 12 and 15, 16 carbon atoms, respectively. Bold is best and underline second best.}
    \vspace{-10pt}
    \label{tab:error_table}
\end{table}

\subsection{Performance on pure graphs}
\label{sec:sec-pure-graph}
\vspace{-5pt}
In this section, we evaluate the performance of our MFNs models in graph-level prediction tasks using GCN layers for the matrix construction. Detailed of the datasets and hyperparameters of the models can be found in the Appendix~\ref{sec:appendix-numerical}.  

\textbf{ZINC.}
We use the default dataset splits for ZINC, aligned with the leaderboard baseline settings, with approximately 500K parameters set. Table~\ref{tab:zinc_results} shows that MFN surpasses previous architectures, demonstrating the utility of learning various operators, even on pure graphs.
\begin{table}[h!t]
\centering
\caption{Results on ZINC with the MAE and number of parameters used, where the best results are in bold. Baselines are taken from~\citep{yang2023better} and model citations are in~\ref{sec:baseline-zinc}.}
\label{tab:zinc_results}
\vspace{-10pt}
\resizebox{0.9\linewidth}{!}{%
    \begin{tabular}{l|cccccccc}
        \toprule
        Method & GCN & GAT & MPNN & GT & SAN & Graphormer & PDF & MFN (GCN) \\
        \midrule
        $\textrm{MAE} $ & 0.367$\pm$0.011 & 0.384$\pm$0.007 & 0.145$\pm$0.007 & 0.226$\pm$0.014 & 0.139$\pm$0.006 & 0.122$\pm$0.006 & 0.066$\pm$0.004 & \textbf{0.063$\pm$0.002}\\
        $\#\textrm{para}$ & 505k & 531k & 481k & NA & 509k & 489k & 472k & 512k \\
        \bottomrule
    \end{tabular}
}
\end{table}

\textbf{TU Datasests.}
We test our model on five TUDataset datasets involving both bioinformatics datasets (MUTAG, ENZYMES, PTC MR, and PROTEINS) and a social network dataset (IMDB-B). 
To ensure a fair comparison with baselines, we follow the standard 10-fold cross-validation and dataset split in table~\ref{tab:tu_results}.
\vspace{-4pt}
\begin{table*}[h!]
	\centering
	\caption{Results on TUDataset (Higher is better). Bold is best, and underlined second best within $\pm 0.5\%$. Baselines are taken from~\citep{yang2023better} and model citations are in~\ref{sec:baseline-tu}}
	\label{tab:tu_results}
	\vspace{-5pt}
	\resizebox{0.70\textwidth}{!}{%
		\begin{tabular}{l|cccccc}
			\toprule
			Method    & MUTAG & ENZYMES & PTC\_MR & PROTEINS & IMDB-B \\
			\midrule
			GK & 81.52$\pm$2.11 & 32.70$\pm$1.20 & 55.65$\pm$0.5 & 71.39$\pm$0.3 & - \\
			RW & 79.11$\pm$2.1 & 24.16$\pm$1.64 & 55.91$\pm$0.3 & 59.57$\pm$0.1 & - \\
			PK & 76.0$\pm$2.7 & - & 59.5$\pm$2.4 & 73.68$\pm$0.7 & - \\
			AWE & 87.87$\pm$9.76 & 35.77$\pm$5.93 & - & - & 74.45$\pm$5.80 \\
			\midrule
			PSCN & 88.95$\pm$4.4 & - & 62.29$\pm$5.7 & 75$\pm$2.5 & 71$\pm$2.3 \\
			ECC & 76.11 & 45.67 & - & - & - \\
			DGK & 87.44$\pm$2.72 & 53.43$\pm$0.91 & 60.08$\pm$2.6 & 75.68$\pm$0.5 & 66.96$\pm$0.6 \\
			GraphSAGE & 85.1$\pm$7.6 & 58.2$\pm$6.0 & - & - & 72.3$\pm$5.3 \\
			CapsGNN & 88.67$\pm$6.88 & 54.67$\pm$5.67 & - & \underline{76.2$\pm$3.6} & 73.1$\pm$4.8 \\
			GIN & 89.4$\pm$5.6 & - & 64.6$\pm$7.0 & 76.2$\pm$2.8 & \underline{75.1$\pm$5.1} \\
			$k$-GNN & 86.1 & - & 60.9 & 75.5 & 74.2 \\
			IGN & 83.89$\pm$12.95 & - & 58.53$\pm$6.86 & \underline{76.58$\pm$5.49} & 72.0$\pm$5.54 \\
			PPGNN & \underline{90.55$\pm$8.7} & - & 66.17$\pm$6.54 & \textbf{77.20$\pm$4.73} & 73.0$\pm$5.77 \\
			GCN$^2$ & 89.39$\pm$1.60 & - & 66.84$\pm$1.79 & 71.71$\pm$1.04 & 74.80$\pm$2.01 \\
			
			PDF  & 89.91$\pm$4.35 & \textbf{73.50}$\pm$\textbf{6.39} & \underline{68.36$\pm$8.38} & \underline{76.28$\pm$5.1} & \textbf{75.60}$\pm$\textbf{2.69} \\
             \midrule
                MFN (GCN) &\textbf{91.5$\pm$7.35} &  \underline{72.9$\pm$7.55}&\textbf{68.9$\pm$8.09}& \underline{76.18$\pm$4.07}& 74.1$\pm$1.04\\
			\bottomrule
		\end{tabular}
	}
	\vspace{-10pt}
\end{table*}

\section{Conclusion}
\vspace{-8pt}
We have introduced Matrix Function Networks (MFNs), an architecture designed to address the limitations of existing GNNs and MPNNs in modeling non-local many-body interactions. Utilizing a resolvent expansion, MFNs achieve potentially linear scaling with respect to system size, offering a computationally efficient solution. Our evaluations indicate state-of-the-art performance on ZINC and TU graph datasets without human-designed features to capture the global graph topology. We also demonstrate that our architecture is capable of modeling the complex non-local interactions of cumulene quantum systems. Future work could focus on extending MFNs to other complex systems, further validating its adaptability and efficiency, and exploring its interpretability.

\subsubsection*{Reproducibility Statement}
To ensure reproducibility and completeness, we include detailed descriptions of the model used, hyperparameters, and data sets in the Appendix. The ZINC and TU datasets are publicly available. We also attach our cumulene datasets in a supplementary. The code is made accessible to the reviewers and will be made public upon publication.


\subsubsection*{Acknowledgments}
CO's work was supported by the NSERC Discovery Grant IDGR019381 and the NFRF Exploration Grant GR022937. 
LLS acknowledges support from the EPSRC Syntech CDT with grant reference EP/S024220/1. 
Computational resources were provided by the Cambridge Service for Data Driven Discovery (CSD3), which was accessed through the University of Cambridge EPSRC Core Equipment Award EP/X034712/1. The authors would also like to thank Rokas Elijošius for useful discussions. 

\subsubsection*{Ethics Statement}
FAF is employed by AstraZeneca at time of publication; however, none of the work presented in this manuscript was conducted at or influenced by this affiliation.

\bibliography{iclr2024_conference}

\begin{thebibliography}{89}
\providecommand{\natexlab}[1]{#1}
\providecommand{\url}[1]{\texttt{#1}}
\expandafter\ifx\csname urlstyle\endcsname\relax
  \providecommand{\doi}[1]{doi: #1}\else
  \providecommand{\doi}{doi: \begingroup \urlstyle{rm}\Url}\fi

\bibitem[Anderson et~al.(2019)Anderson, Hy, and
  Kondor]{Anderson2019CormorantCM}
Brandon Anderson, Truong~Son Hy, and Risi Kondor.
\newblock Cormorant: Covariant molecular neural networks.
\newblock In H.~Wallach, H.~Larochelle, A.~Beygelzimer, F.~AlcheBuc, E.~Fox,
  and R.~Garnett (eds.), \emph{Advances in Neural Information Processing
  Systems}, volume~32. Curran Associates, Inc., 2019.
\newblock URL
  \url{https://proceedings.neurips.cc/paper/2019/file/03573b32b2746e6e8ca98b9123f2249b-Paper.pdf}.

\bibitem[Ba et~al.(2016)Ba, Kiros, and Hinton]{ba2016layer}
Jimmy~Lei Ba, Jamie~Ryan Kiros, and Geoffrey~E Hinton.
\newblock Layer normalization.
\newblock \emph{arXiv preprint arXiv:1607.06450}, 2016.

\bibitem[Batatia et~al.(2022{\natexlab{a}})Batatia, Batzner, Kovács,
  Musaelian, Simm, Drautz, Ortner, Kozinsky, and Csányi]{Batatia2022de}
Ilyes Batatia, Simon Batzner, Dávid~Péter Kovács, Albert Musaelian, Gregor
  N.~C. Simm, Ralf Drautz, Christoph Ortner, Boris Kozinsky, and Gábor
  Csányi.
\newblock The design space of e(3)-equivariant atom-centered interatomic
  potentials, 2022{\natexlab{a}}.
\newblock URL \url{https://arxiv.org/abs/2205.06643}.

\bibitem[Batatia et~al.(2022{\natexlab{b}})Batatia, Kovács, Simm, Ortner, and
  Csányi]{Batatia2022mace}
Ilyes Batatia, Dávid~Péter Kovács, Gregor N.~C. Simm, Christoph Ortner, and
  Gábor Csányi.
\newblock Mace: Higher order equivariant message passing neural networks for
  fast and accurate force fields, 2022{\natexlab{b}}.
\newblock URL \url{https://arxiv.org/abs/2206.07697}.

\bibitem[Batatia et~al.(2023)Batatia, Geiger, Munoz, Smidt, Silberman, and
  Ortner]{batatia2023general}
Ilyes Batatia, Mario Geiger, Jose Munoz, Tess Smidt, Lior Silberman, and
  Christoph Ortner.
\newblock A general framework for equivariant neural networks on reductive lie
  groups, 2023.

\bibitem[Battaglia et~al.(2018)Battaglia, Hamrick, Bapst, Sanchez-Gonzalez,
  Zambaldi, Malinowski, Tacchetti, Raposo, Santoro, Faulkner,
  et~al.]{battaglia2018relational}
Peter~W Battaglia, Jessica~B Hamrick, Victor Bapst, Alvaro Sanchez-Gonzalez,
  Vinicius Zambaldi, Mateusz Malinowski, Andrea Tacchetti, David Raposo, Adam
  Santoro, Ryan Faulkner, et~al.
\newblock Relational inductive biases, deep learning, and graph networks.
\newblock \emph{arXiv preprint arXiv:1806.01261}, 2018.

\bibitem[Batzner et~al.(2022)Batzner, Musaelian, Sun, Geiger, Mailoa,
  Kornbluth, Molinari, Smidt, and Kozinsky]{nequip}
Simon Batzner, Albert Musaelian, Lixin Sun, Mario Geiger, Jonathan~P. Mailoa,
  Mordechai Kornbluth, Nicola Molinari, Tess~E. Smidt, and Boris Kozinsky.
\newblock E(3)-equivariant graph neural networks for data-efficient and
  accurate interatomic potentials.
\newblock \emph{Nature Communications}, 13\penalty0 (1):\penalty0 2453, 2022.

\bibitem[Behler(2021)]{behler2021four_gen_rev}
Jörg Behler.
\newblock Four generations of high-dimensional neural network potentials.
\newblock \emph{Chemical Reviews}, 121\penalty0 (16):\penalty0 10037--10072,
  2021.

\bibitem[Bianchi et~al.(2021{\natexlab{a}})Bianchi, Grattarola, Livi, and
  Alippi]{ARMA}
Filippo~Maria Bianchi, Daniele Grattarola, Lorenzo Livi, and Cesare Alippi.
\newblock Graph neural networks with convolutional {ARMA} filters.
\newblock \emph{IEEE Transactions on Pattern Analysis and Machine
  Intelligence}, 2021{\natexlab{a}}.

\bibitem[Bianchi et~al.(2021{\natexlab{b}})Bianchi, Grattarola, Livi, and
  Alippi]{bianchi2021graph}
Filippo~Maria Bianchi, Daniele Grattarola, Lorenzo Livi, and Cesare Alippi.
\newblock Graph neural networks with convolutional arma filters.
\newblock \emph{IEEE transactions on pattern analysis and machine
  intelligence}, 44\penalty0 (7):\penalty0 3496--3507, 2021{\natexlab{b}}.

\bibitem[Brandstetter et~al.(2022)Brandstetter, Hesselink, van~der Pol,
  Bekkers, and Welling]{brandstetter2022geometric}
Johannes Brandstetter, Rob Hesselink, Elise van~der Pol, Erik~J Bekkers, and
  Max Welling.
\newblock Geometric and physical quantities improve e(3) equivariant message
  passing, 2022.

\bibitem[Bronstein et~al.(2021)Bronstein, Bruna, Cohen, and
  Veli\v{c}kovi\'c]{Bronstein:2021mdi}
Michael~M. Bronstein, Joan Bruna, Taco Cohen, and Petar Veli\v{c}kovi\'c.
\newblock {Geometric Deep Learning: Grids, Groups, Graphs, Geodesics, and
  Gauges}, 4 2021.

\bibitem[Chien et~al.(2021)Chien, Peng, Li, and Milenkovic]{GPRGNN}
Eli Chien, Jianhao Peng, Pan Li, and Olgica Milenkovic.
\newblock Adaptive universal generalized pagerank graph neural network.
\newblock In \emph{International Conference on Learning Representations}, 2021.

\bibitem[Cho et~al.(2014)Cho, Van~Merri{\"e}nboer, Gulcehre, Bahdanau,
  Bougares, Schwenk, and Bengio]{cho2014properties}
Kyunghyun Cho, Bart Van~Merri{\"e}nboer, Caglar Gulcehre, Dzmitry Bahdanau,
  Fethi Bougares, Holger Schwenk, and Yoshua Bengio.
\newblock On the properties of neural machine translation: Encoder-decoder
  approaches.
\newblock In \emph{Proceedings of the 2014 Conference on Empirical Methods in
  Natural Language Processing (EMNLP)}, pp.\  1724--1734, 2014.

\bibitem[Cohen \& Welling(2016)Cohen and Welling]{CohenSteerable2016}
Taco~S. Cohen and Max Welling.
\newblock Steerable cnns.
\newblock \emph{ICLR 2017}, 2016.
\newblock \doi{10.48550/ARXIV.1612.08498}.
\newblock URL \url{https://arxiv.org/abs/1612.08498}.

\bibitem[Cohen et~al.(2018)Cohen, Geiger, Köhler, and
  Welling]{s.2018spherical}
Taco~S. Cohen, Mario Geiger, Jonas Köhler, and Max Welling.
\newblock Spherical {CNN}s.
\newblock In \emph{International Conference on Learning Representations}, 2018.
\newblock URL \url{https://openreview.net/forum?id=Hkbd5xZRb}.

\bibitem[Combes \& Thomas(1973)Combes and Thomas]{Combes1973}
J.~M. Combes and L.~Thomas.
\newblock Asymptotic behaviour of eigenfunctions for multiparticle
  schr\"{o}dinger operators.
\newblock \emph{Communications in Mathematical Physics}, 34\penalty0
  (4):\penalty0 251--270, December 1973.
\newblock \doi{10.1007/bf01646473}.
\newblock URL \url{https://doi.org/10.1007/bf01646473}.

\bibitem[Darby et~al.(2023)Darby, Kov\'acs, Batatia, Caro, Hart, Ortner, and
  Cs\'anyi]{TraceDarby2023}
James~P. Darby, D\'avid~P. Kov\'acs, Ilyes Batatia, Miguel~A. Caro, Gus L.~W.
  Hart, Christoph Ortner, and G\'abor Cs\'anyi.
\newblock Tensor-reduced atomic density representations.
\newblock \emph{Phys. Rev. Lett.}, 131:\penalty0 028001, Jul 2023.
\newblock \doi{10.1103/PhysRevLett.131.028001}.
\newblock URL \url{https://link.aps.org/doi/10.1103/PhysRevLett.131.028001}.

\bibitem[Davis(2006)]{davis2006}
Timothy~A. Davis.
\newblock \emph{Direct Methods for Sparse Linear Systems}.
\newblock Society for Industrial and Applied Mathematics, 2006.
\newblock \doi{10.1137/1.9780898718881}.
\newblock URL \url{https://epubs.siam.org/doi/abs/10.1137/1.9780898718881}.

\bibitem[de~Haan et~al.(2020)de~Haan, Cohen, and Welling]{de2020natural}
Pim de~Haan, Taco~S Cohen, and Max Welling.
\newblock Natural graph networks.
\newblock \emph{Advances in Neural Information Processing Systems},
  33:\penalty0 3636--3646, 2020.

\bibitem[Defferrard et~al.(2016{\natexlab{a}})Defferrard, Bresson, and
  Vandergheynst]{ChebyConv}
Micha{\"{e}}l Defferrard, Xavier Bresson, and Pierre Vandergheynst.
\newblock Convolutional neural networks on graphs with fast localized spectral
  filtering.
\newblock In \emph{Advances in Neural Information Processing Systems}, pp.\
  3837--3845, 2016{\natexlab{a}}.

\bibitem[Defferrard et~al.(2016{\natexlab{b}})Defferrard, Bresson, and
  Vandergheynst]{defferrard2016convolutional}
Micha{\"e}l Defferrard, Xavier Bresson, and Pierre Vandergheynst.
\newblock Convolutional neural networks on graphs with fast localized spectral
  filtering.
\newblock \emph{Advances in neural information processing systems}, 29,
  2016{\natexlab{b}}.

\bibitem[Di~Giovanni et~al.(2023)Di~Giovanni, Giusti, Barbero, Luise, Lio, and
  Bronstein]{di2023over}
Francesco Di~Giovanni, Lorenzo Giusti, Federico Barbero, Giulia Luise, Pietro
  Lio, and Michael Bronstein.
\newblock On over-squashing in message passing neural networks: The impact of
  width, depth, and topology.
\newblock \emph{arXiv preprint arXiv:2302.02941}, 2023.

\bibitem[Drautz(2020)]{Drautz2020-tensors}
Ralf Drautz.
\newblock Atomic cluster expansion of scalar, vectorial, and tensorial
  properties including magnetism and charge transfer.
\newblock \emph{Phys. Rev. B}, 102:\penalty0 024104, Jul 2020.
\newblock \doi{10.1103/PhysRevB.102.024104}.
\newblock URL \url{https://link.aps.org/doi/10.1103/PhysRevB.102.024104}.

\bibitem[Dwivedi et~al.(2020)Dwivedi, Joshi, Laurent, Bengio, and
  Bresson]{dwivedi2020benchmarking}
Vijay~Prakash Dwivedi, Chaitanya~K Joshi, Thomas Laurent, Yoshua Bengio, and
  Xavier Bresson.
\newblock Benchmarking graph neural networks.
\newblock \emph{arXiv preprint arXiv:2003.00982}, 2020.

\bibitem[Elman(1990)]{elman1990finding}
Jeffrey~L Elman.
\newblock Finding structure in time.
\newblock \emph{Cognitive Science}, 14\penalty0 (2):\penalty0 179--211, 1990.

\bibitem[Etter(2020)]{etter2020incomplete}
Simon Etter.
\newblock Incomplete selected inversion for linear-scaling electronic structure
  calculations, 2020.

\bibitem[Frank et~al.(2022)Frank, Unke, and M{\"u}ller]{frank2022so3krates}
Thorben Frank, Oliver Unke, and Klaus-Robert M{\"u}ller.
\newblock So3krates: Equivariant attention for interactions on arbitrary
  length-scales in molecular systems.
\newblock \emph{Advances in Neural Information Processing Systems},
  35:\penalty0 29400--29413, 2022.

\bibitem[Gao \& Remsing(2022)Gao and Remsing]{gao2022self_consistent_nn}
Ang Gao and Richard~C Remsing.
\newblock Self-consistent determination of long-range electrostatics in neural
  network potentials.
\newblock \emph{Nature communications}, 13\penalty0 (1):\penalty0 1572, 2022.

\bibitem[Gasteiger et~al.(2018)Gasteiger, Bojchevski, and
  G{\"u}nnemann]{gasteiger2018predict}
Johannes Gasteiger, Aleksandar Bojchevski, and Stephan G{\"u}nnemann.
\newblock Predict then propagate: Graph neural networks meet personalized
  pagerank.
\newblock \emph{arXiv preprint arXiv:1810.05997}, 2018.

\bibitem[Gilmer et~al.(2017)Gilmer, Schoenholz, Riley, Vinyals, and
  Dahl]{gilmer2017neural}
Justin Gilmer, Samuel~S Schoenholz, Patrick~F Riley, Oriol Vinyals, and
  George~E Dahl.
\newblock Neural message passing for quantum chemistry.
\newblock In \emph{Proceedings of the 34th International Conference on Machine
  Learning-Volume 70}, pp.\  1263--1272. JMLR. org, 2017.

\bibitem[Goedecker(1999)]{RevModPhys.71.1085}
Stefan Goedecker.
\newblock Linear scaling electronic structure methods.
\newblock \emph{Rev. Mod. Phys.}, 71:\penalty0 1085--1123, Jul 1999.
\newblock \doi{10.1103/RevModPhys.71.1085}.
\newblock URL \url{https://link.aps.org/doi/10.1103/RevModPhys.71.1085}.

\bibitem[Graves(2013)]{graves2013generating}
Alex Graves.
\newblock Generating sequences with recurrent neural networks.
\newblock In \emph{arXiv preprint arXiv:1308.0850}, 2013.

\bibitem[Grisafi \& Ceriotti(2019)Grisafi and Ceriotti]{Grisafi2019LODE1}
Andrea Grisafi and Michele Ceriotti.
\newblock {Incorporating long-range physics in atomic-scale machine learning}.
\newblock \emph{The Journal of Chemical Physics}, 151\penalty0 (20), 11 2019.
\newblock ISSN 0021-9606.
\newblock \doi{10.1063/1.5128375}.
\newblock URL \url{https://doi.org/10.1063/1.5128375}.
\newblock 204105.

\bibitem[Gu et~al.(2023)Gu, Zhouyin, Pandey, Zhang, Zhang, and
  Weinan]{Gu2023-om}
Qiangqiang Gu, Zhanghao Zhouyin, Shishir~Kumar Pandey, Peng Zhang, Linfeng
  Zhang, and E~Weinan.
\newblock {DeePTB}: A deep learning-based tight-binding approach with $ab$
  $initio$ accuracy.
\newblock \emph{arXiv}, July 2023.

\bibitem[Hamilton et~al.(2017)Hamilton, Ying, and
  Leskovec]{hamilton2017inductive}
Will Hamilton, Zhitao Ying, and Jure Leskovec.
\newblock Inductive representation learning on large graphs.
\newblock In \emph{Advances in Neural Information Processing Systems}, pp.\
  1024--1034, 2017.

\bibitem[He et~al.(2021)He, Wei, Xu, et~al.]{he2021bernnet}
Mingguo He, Zhewei Wei, Hongteng Xu, et~al.
\newblock Bernnet: Learning arbitrary graph spectral filters via bernstein
  approximation.
\newblock \emph{Advances in Neural Information Processing Systems},
  34:\penalty0 14239--14251, 2021.

\bibitem[Hegde \& Bowen(2017)Hegde and Bowen]{Hegde2017-ay}
Ganesh Hegde and R~Chris Bowen.
\newblock Machine-learned approximations to density functional theory
  hamiltonians.
\newblock \emph{Sci. Rep.}, 7:\penalty0 42669, February 2017.

\bibitem[Higham(2008)]{higham2008functions}
Nicholas~J Higham.
\newblock \emph{Functions of matrices: theory and computation}.
\newblock SIAM, 2008.

\bibitem[Hochreiter \& Schmidhuber(1997)Hochreiter and
  Schmidhuber]{hochreiter1997long}
Sepp Hochreiter and J{\"u}rgen Schmidhuber.
\newblock Long short-term memory.
\newblock In \emph{Neural computation}, volume 9,8, pp.\  1735--1780. MIT
  Press, 1997.

\bibitem[Hu et~al.(2019)Hu, Jin, Yang, and Zhang]{hu2019deep}
Zhixuan Hu, Li~Jin, Yi~Yang, and Ronghang Zhang.
\newblock Deep graph pose: a semi-supervised deep graphical model for improved
  animal pose tracking.
\newblock In \emph{Proceedings of the 27th ACM International Conference on
  Multimedia}, pp.\  1365--1374. ACM, 2019.

\bibitem[Huguenin-Dumittan et~al.(2023)Huguenin-Dumittan, Loche, Haoran, and
  Ceriotti]{huguenindumittan2023physicsinspired}
Kevin~K. Huguenin-Dumittan, Philip Loche, Ni~Haoran, and Michele Ceriotti.
\newblock Physics-inspired equivariant descriptors of non-bonded interactions,
  2023.

\bibitem[Ioffe \& Szegedy(2015)Ioffe and Szegedy]{ioffe2015batch}
Sergey Ioffe and Christian Szegedy.
\newblock Batch normalization: Accelerating deep network training by reducing
  internal covariate shift.
\newblock In \emph{International conference on machine learning}, pp.\
  448--456. pmlr, 2015.

\bibitem[Irwin \& Shoichet(2004)Irwin and Shoichet]{Irwin2004}
John~J. Irwin and Brian~K. Shoichet.
\newblock {ZINC} - a free database of commercially available compounds for
  virtual screening.
\newblock \emph{Journal of Chemical Information and Modeling}, 45\penalty0
  (1):\penalty0 177--182, December 2004.
\newblock \doi{10.1021/ci049714+}.
\newblock URL \url{https://doi.org/10.1021/ci049714+}.

\bibitem[Ivanov \& Burnaev(2018)Ivanov and Burnaev]{pmlr-v80-ivanov18a}
Sergey Ivanov and Evgeny Burnaev.
\newblock Anonymous walk embeddings.
\newblock In Jennifer Dy and Andreas Krause (eds.), \emph{Proceedings of the
  35th International Conference on Machine Learning}, volume~80 of
  \emph{Proceedings of Machine Learning Research}, pp.\  2191--2200,
  Stockholmsmässan, Stockholm Sweden, 10--15 Jul 2018. PMLR.
\newblock URL \url{http://proceedings.mlr.press/v80/ivanov18a.html}.

\bibitem[Jaegle et~al.(2022)Jaegle, Borgeaud, Alayrac, Doersch, Ionescu, Ding,
  Koppula, Zoran, Brock, Shelhamer, Hénaff, Botvinick, Zisserman, Vinyals, and
  Carreira]{jaegle2022perceiver}
Andrew Jaegle, Sebastian Borgeaud, Jean-Baptiste Alayrac, Carl Doersch, Catalin
  Ionescu, David Ding, Skanda Koppula, Daniel Zoran, Andrew Brock, Evan
  Shelhamer, Olivier Hénaff, Matthew~M. Botvinick, Andrew Zisserman, Oriol
  Vinyals, and Joāo Carreira.
\newblock Perceiver io: A general architecture for structured inputs and
  outputs, 2022.

\bibitem[Johnson \& Newman(1980)Johnson and Newman]{JOHNSON198096}
Charles~R Johnson and Morris Newman.
\newblock A note on cospectral graphs.
\newblock \emph{Journal of Combinatorial Theory, Series B}, 28\penalty0
  (1):\penalty0 96--103, 1980.
\newblock ISSN 0095-8956.
\newblock \doi{https://doi.org/10.1016/0095-8956(80)90058-1}.
\newblock URL
  \url{https://www.sciencedirect.com/science/article/pii/0095895680900581}.

\bibitem[Joshi et~al.(2023)Joshi, Bodnar, Mathis, Cohen, and
  Liò]{joshi2023expressive}
Chaitanya~K. Joshi, Cristian Bodnar, Simon~V. Mathis, Taco Cohen, and Pietro
  Liò.
\newblock On the expressive power of geometric graph neural networks.
\newblock In \emph{International Conference on Machine Learning}, 2023.

\bibitem[Kipf \& Welling(2017{\natexlab{a}})Kipf and Welling]{kipf2016semi}
Thomas~N Kipf and Max Welling.
\newblock Semi-supervised classification with graph convolutional networks.
\newblock In \emph{Proceedings of the International Conference on Learning
  Representations (ICLR)}, 2017{\natexlab{a}}.

\bibitem[Kipf \& Welling(2017{\natexlab{b}})Kipf and Welling]{kipf2017semi}
Thomas~N. Kipf and Max Welling.
\newblock Semi-supervised classification with graph convolutional networks.
\newblock In \emph{International Conference on Learning Representations
  (ICLR)}, 2017{\natexlab{b}}.

\bibitem[Kondor \& Trivedi(2018)Kondor and Trivedi]{pmlr-v80-kondor18a}
Risi Kondor and Shubhendu Trivedi.
\newblock On the generalization of equivariance and convolution in neural
  networks to the action of compact groups.
\newblock In Jennifer Dy and Andreas Krause (eds.), \emph{Proceedings of the
  35th International Conference on Machine Learning}, volume~80 of
  \emph{Proceedings of Machine Learning Research}, pp.\  2747--2755. PMLR,
  10--15 Jul 2018.
\newblock URL \url{https://proceedings.mlr.press/v80/kondor18a.html}.

\bibitem[Kosmala et~al.(2023)Kosmala, Gasteiger, Gao, and
  Günnemann]{kosmala2023ewaldbased}
Arthur Kosmala, Johannes Gasteiger, Nicholas Gao, and Stephan Günnemann.
\newblock Ewald-based long-range message passing for molecular graphs, 2023.

\bibitem[Kovacs et~al.(2023)Kovacs, Batatia, Arany, and
  Csanyi]{kovacsEvaluationMACEForce2023}
David~Peter Kovacs, Ilyes Batatia, Eszter~Sara Arany, and Gabor Csanyi.
\newblock Evaluation of the {{MACE Force Field Architecture}}: From {{Medicinal
  Chemistry}} to {{Materials Science}}.
\newblock \emph{The Journal of Chemical Physics}, 159\penalty0 (4):\penalty0
  044118, July 2023.
\newblock ISSN 0021-9606, 1089-7690.
\newblock \doi{10.1063/5.0155322}.

\bibitem[Kreuzer et~al.(2021)Kreuzer, Beaini, Hamilton, L{\'e}tourneau, and
  Tossou]{kreuzer2021rethinking}
Devin Kreuzer, Dominique Beaini, Will Hamilton, Vincent L{\'e}tourneau, and
  Prudencio Tossou.
\newblock Rethinking graph transformers with spectral attention.
\newblock \emph{Advances in Neural Information Processing Systems},
  34:\penalty0 21618--21629, 2021.

\bibitem[LeCun et~al.(1989)LeCun, Boser, Denker, Henderson, Howard, Hubbard,
  and Jackel]{CNNlecun}
Y.~LeCun, B.~Boser, J.~S. Denker, D.~Henderson, R.~E. Howard, W.~Hubbard, and
  L.~D. Jackel.
\newblock Backpropagation applied to handwritten zip code recognition.
\newblock \emph{Neural Computation}, 1\penalty0 (4):\penalty0 541--551, 1989.
\newblock \doi{10.1162/neco.1989.1.4.541}.

\bibitem[Lin et~al.(2009)Lin, Lu, Ying, Car, and E]{pexsi_CMS2009}
L.~Lin, J.~Lu, L.~Ying, R.~Car, and W.~E.
\newblock Fast algorithm for extracting the diagonal of the inverse matrix with
  application to the electronic structure analysis of metallic systems.
\newblock \emph{Comm. Math. Sci.}, 7:\penalty0 755, 2009.

\bibitem[Lin et~al.(2013)Lin, Chen, Yang, and He]{pexsi_JCPM2013}
L.~Lin, M.~Chen, C.~Yang, and L.~He.
\newblock Accelerating atomic orbital-based electronic structure calculation
  via pole expansion and selected inversion.
\newblock \emph{J. Phys. Condens. Matter}, 25:\penalty0 295501, 2013.

\bibitem[Maron et~al.(2018)Maron, Ben-Hamu, Shamir, and
  Lipman]{maron2018invariant}
Haggai Maron, Heli Ben-Hamu, Nadav Shamir, and Yaron Lipman.
\newblock Invariant and equivariant graph networks.
\newblock \emph{arXiv preprint arXiv:1812.09902}, 2018.

\bibitem[Maron et~al.(2019)Maron, Ben-Hamu, Serviansky, and
  Lipman]{maron2019provably}
Haggai Maron, Heli Ben-Hamu, Hadar Serviansky, and Yaron Lipman.
\newblock Provably powerful graph networks.
\newblock In \emph{Proceedings of the 33rd International Conference on Neural
  Information Processing Systems}, pp.\  2156--2167, 2019.

\bibitem[Morris et~al.(2019)Morris, Ritzert, Fey, Hamilton, Lenssen, Rattan,
  and Grohe]{morris2019weisfeiler}
Christopher Morris, Martin Ritzert, Matthias Fey, William~L Hamilton, Jan~Eric
  Lenssen, Gaurav Rattan, and Martin Grohe.
\newblock Weisfeiler and leman go neural: Higher-order graph neural networks.
\newblock In \emph{Proceedings of the AAAI Conference on Artificial
  Intelligence}, volume~33, pp.\  4602--4609, 2019.

\bibitem[Neumann et~al.(2016)Neumann, Garnett, Bauckhage, and
  Kersting]{neumann2016propagation}
Marion Neumann, Roman Garnett, Christian Bauckhage, and Kristian Kersting.
\newblock Propagation kernels: efficient graph kernels from propagated
  information.
\newblock \emph{Machine Learning}, 102\penalty0 (2):\penalty0 209--245, 2016.

\bibitem[Niepert et~al.(2016)Niepert, Ahmed, and Kutzkov]{niepert2016learning}
Mathias Niepert, Mohamed Ahmed, and Konstantin Kutzkov.
\newblock Learning convolutional neural networks for graphs.
\newblock In \emph{International Conference on Machine Learning}, pp.\
  2014--2023, 2016.

\bibitem[Nigam et~al.(2022)Nigam, Willatt, and
  Ceriotti]{Nigam2022-hamiltonians}
Jigyasa Nigam, Michael~J. Willatt, and Michele Ceriotti.
\newblock Equivariant representations for molecular hamiltonians and n-center
  atomic-scale properties.
\newblock \emph{J. Chem. Phys.}, 156, 2022.

\bibitem[Parsaeifard et~al.(2021)Parsaeifard, De, Christensen, Faber, Kocer,
  De, Behler, von Lilienfeld, and Goedecker]{OMFPs2021}
Behnam Parsaeifard, Deb~Sankar De, Anders~S Christensen, Felix~A Faber, Emir
  Kocer, Sandip De, Jörg Behler, O~Anatole von Lilienfeld, and Stefan
  Goedecker.
\newblock An assessment of the structural resolution of various fingerprints
  commonly used in machine learning.
\newblock \emph{Machine Learning: Science and Technology}, 2\penalty0
  (1):\penalty0 015018, apr 2021.
\newblock \doi{10.1088/2632-2153/abb212}.
\newblock URL \url{https://dx.doi.org/10.1088/2632-2153/abb212}.

\bibitem[Satorras et~al.(2021)Satorras, Hoogeboom, and
  Welling]{Welling2021EGNN}
Victor~Garcia Satorras, Emiel Hoogeboom, and Max Welling.
\newblock E(n) equivariant graph neural networks, 2021.
\newblock URL \url{https://arxiv.org/abs/2102.09844}.

\bibitem[Sch{\"u}tt et~al.(2019)Sch{\"u}tt, Gastegger, Tkatchenko, M{\"u}ller,
  and Maurer]{Schutt2019-um}
K~T Sch{\"u}tt, M~Gastegger, A~Tkatchenko, K-R M{\"u}ller, and R~J Maurer.
\newblock Unifying machine learning and quantum chemistry with a deep neural
  network for molecular wavefunctions.
\newblock \emph{Nat. Commun.}, 10\penalty0 (1):\penalty0 5024, November 2019.

\bibitem[Senior et~al.(2018)Senior, Evans, Jumper, Kirkpatrick, Sifre, Green,
  Qin, Zidek, Nelson, Bridgland, et~al.]{deepmind2018protein}
Andrew~W Senior, Richard Evans, John Jumper, Jeff Kirkpatrick, Laurent Sifre,
  Tim Green, Chongli Qin, Augustin Zidek, Alexander W~R Nelson, Alex Bridgland,
  et~al.
\newblock Protein folding using distance maps and deep learning.
\newblock \emph{bioRxiv}, 2018.

\bibitem[Shaw et~al.(2018)Shaw, Uszkoreit, and Vaswani]{shaw2018selfattention}
Peter Shaw, Jakob Uszkoreit, and Ashish Vaswani.
\newblock Self-attention with relative position representations, 2018.

\bibitem[Shervashidze et~al.(2009)Shervashidze, Vishwanathan, Petri, Mehlhorn,
  and Borgwardt]{shervashidze2009efficient}
Nino Shervashidze, SVN Vishwanathan, Tobias Petri, Kurt Mehlhorn, and Karsten
  Borgwardt.
\newblock Efficient graphlet kernels for large graph comparison.
\newblock In \emph{Artificial Intelligence and Statistics}, pp.\  488--495,
  2009.

\bibitem[Simonovsky \& Komodakis(2017)Simonovsky and
  Komodakis]{simonovsky2017dynamic}
Martin Simonovsky and Nikos Komodakis.
\newblock Dynamic edge-conditioned filters in convolutional neural networks on
  graphs.
\newblock In \emph{Proceedings of the IEEE conference on computer vision and
  pattern recognition}, pp.\  3693--3702, 2017.

\bibitem[Unke et~al.(2021{\natexlab{a}})Unke, Bogojeski, Gastegger, Geiger,
  Smidt, and M\"{u}ller]{Unke-sg-2021}
O.~T. Unke, M.~Bogojeski, M.~Gastegger, M.~Geiger, T.~Smidt, and K.-R.
  M\"{u}ller.
\newblock {SE} (3)-equivariant prediction of molecular wavefunctions and
  electronic densities, 2021{\natexlab{a}}.
\newblock 35th Conference on Neural Information Processing Systems (NeurIPS
  2021).

\bibitem[Unke \& Meuwly(2019)Unke and Meuwly]{Unke2019PhysNet:Charges}
Oliver~T. Unke and Markus Meuwly.
\newblock {PhysNet: A Neural Network for Predicting Energies, Forces, Dipole
  Moments, and Partial Charges}.
\newblock \emph{Journal of Chemical Theory and Computation}, 15\penalty0
  (6):\penalty0 3678--3693, 6 2019.
\newblock ISSN 15499626.
\newblock \doi{10.1021/acs.jctc.9b00181}.

\bibitem[Unke et~al.(2021{\natexlab{b}})Unke, Chmiela, Gastegger, Sch\"{u}tt,
  Sauceda, and M\"{u}ller]{Unke2021}
Oliver~T. Unke, Stefan Chmiela, Michael Gastegger, Kristof~T. Sch\"{u}tt,
  Huziel~E. Sauceda, and Klaus-Robert M\"{u}ller.
\newblock {SpookyNet}: Learning force fields with electronic degrees of freedom
  and nonlocal effects.
\newblock \emph{Nature Communications}, 12\penalty0 (1), December
  2021{\natexlab{b}}.
\newblock \doi{10.1038/s41467-021-27504-0}.
\newblock URL \url{https://doi.org/10.1038/s41467-021-27504-0}.

\bibitem[Unke et~al.(2021{\natexlab{c}})Unke, Chmiela, Sauceda, Gastegger,
  Poltavsky, Schütt, Tkatchenko, and Müller]{unke2021machine}
Oliver~T Unke, Stefan Chmiela, Huziel~E Sauceda, Michael Gastegger, Igor
  Poltavsky, Kristof~T Schütt, Alexandre Tkatchenko, and Klaus-Robert Müller.
\newblock Machine learning force fields.
\newblock \emph{Chemical Reviews}, 121\penalty0 (16):\penalty0 10142--10186,
  2021{\natexlab{c}}.

\bibitem[Vaswani et~al.(2017)Vaswani, Shazeer, Parmar, Uszkoreit, Jones, Gomez,
  Kaiser, and Polosukhin]{vaswani2017attention}
Ashish Vaswani, Noam Shazeer, Niki Parmar, Jakob Uszkoreit, Llion Jones,
  Aidan~N Gomez, Lukasz Kaiser, and Illia Polosukhin.
\newblock Attention is all you need.
\newblock \emph{Advances in neural information processing systems},
  30:\penalty0 5998--6008, 2017.

\bibitem[Veli{\v{c}}kovi{\'{c}} et~al.(2018)Veli{\v{c}}kovi{\'{c}}, Cucurull,
  Casanova, Romero, Li{\`{o}}, and Bengio]{velickovic2018graph}
Petar Veli{\v{c}}kovi{\'{c}}, Guillem Cucurull, Arantxa Casanova, Adriana
  Romero, Pietro Li{\`{o}}, and Yoshua Bengio.
\newblock {Graph Attention Networks}.
\newblock \emph{International Conference on Learning Representations}, 2018.
\newblock URL \url{https://openreview.net/forum?id=rJXMpikCZ}.

\bibitem[Veli{\v{c}}kovi{\'c} et~al.(2018)Veli{\v{c}}kovi{\'c}, Cucurull,
  Casanova, Romero, Lio, and Bengio]{velivckovic2017graph}
Petar Veli{\v{c}}kovi{\'c}, Guillem Cucurull, Arantxa Casanova, Adriana Romero,
  Pietro Lio, and Yoshua Bengio.
\newblock Graph attention networks.
\newblock In \emph{Proceedings of the International Conference on Learning
  Representations (ICLR)}, 2018.

\bibitem[Vishwanathan et~al.(2010)Vishwanathan, Schraudolph, Kondor, and
  Borgwardt]{vishwanathan2010graph}
S~Vichy~N Vishwanathan, Nicol~N Schraudolph, Risi Kondor, and Karsten~M
  Borgwardt.
\newblock Graph kernels.
\newblock \emph{Journal of Machine Learning Research}, 11\penalty0
  (Apr):\penalty0 1201--1242, 2010.

\bibitem[Wang \& Zhang(2022)Wang and Zhang]{wang2022powerful}
Xiyuan Wang and Muhan Zhang.
\newblock How powerful are spectral graph neural networks.
\newblock In \emph{International Conference on Machine Learning}, pp.\
  23341--23362. PMLR, 2022.

\bibitem[Wu et~al.(2020)Wu, Pan, Chen, Long, Zhang, and
  Yu]{wu2020comprehensive}
Zonghan Wu, Shirui Pan, Fengwen Chen, Guodong Long, Chengqi Zhang, and Philip~S
  Yu.
\newblock A comprehensive survey on graph neural networks.
\newblock \emph{IEEE Transactions on Neural Networks and Learning Systems},
  32\penalty0 (1):\penalty0 4--24, 2020.

\bibitem[Wu et~al.(2021)Wu, Pan, Chen, Long, Zhang, and Yu]{WUGNN2021}
Zonghan Wu, Shirui Pan, Fengwen Chen, Guodong Long, Chengqi Zhang, and
  Philip~S. Yu.
\newblock A comprehensive survey on graph neural networks.
\newblock \emph{IEEE Transactions on Neural Networks and Learning Systems},
  32\penalty0 (1):\penalty0 4--24, 2021.
\newblock \doi{10.1109/TNNLS.2020.2978386}.

\bibitem[Xinyi \& Chen(2019)Xinyi and Chen]{xinyi2018capsule}
Zhang Xinyi and Lihui Chen.
\newblock Capsule graph neural network.
\newblock In \emph{International Conference on Learning Representations}, 2019.
\newblock URL \url{https://openreview.net/forum?id=Byl8BnRcYm}.

\bibitem[Xu et~al.(2019)Xu, Hu, Leskovec, and Jegelka]{xu2018how}
Keyulu Xu, Weihua Hu, Jure Leskovec, and Stefanie Jegelka.
\newblock How powerful are graph neural networks?
\newblock In \emph{International Conference on Learning Representations}, 2019.
\newblock URL \url{https://openreview.net/forum?id=ryGs6iA5Km}.

\bibitem[Yanardag \& Vishwanathan(2015)Yanardag and
  Vishwanathan]{yanardag2015deep}
Pinar Yanardag and SVN Vishwanathan.
\newblock Deep graph kernels.
\newblock In \emph{Proceedings of the 21th ACM SIGKDD International Conference
  on Knowledge Discovery and Data Mining}, pp.\  1365--1374. ACM, 2015.

\bibitem[Yang et~al.(2023)Yang, Feng, Shen, and Hooi]{yang2023better}
Mingqi Yang, Wenjie Feng, Yanming Shen, and Bryan Hooi.
\newblock Towards better graph representation learning with parameterized
  decomposition and filtering, 2023.

\bibitem[Ying et~al.(2021)Ying, Cai, Luo, Zheng, Ke, He, Shen, and
  Liu]{ying2021transformers}
Chengxuan Ying, Tianle Cai, Shengjie Luo, Shuxin Zheng, Guolin Ke, Di~He,
  Yanming Shen, and Tie-Yan Liu.
\newblock Do transformers really perform bad for graph representation?
\newblock \emph{arXiv preprint arXiv:2106.05234}, 2021.

\bibitem[Zhang et~al.(2022)Zhang, Onat, Dusson, Anand, Maurer, Ortner, and
  Kermode]{2021-acetb1}
Liwei Zhang, Berk Onat, Genevieve Dusson, Gautam Anand, Reinhard~J Maurer,
  Christoph Ortner, and James~R Kermode.
\newblock Equivariant analytical mapping of first principles hamiltonians to
  accurate and transferable materials models.
\newblock \emph{npj Computational Materials}, 8, 2022.
\newblock \doi{https://doi.org/10.1038/s41524-022-00843-2}.
\newblock URL \url{https://arxiv.org/abs/2111.13736}.

\bibitem[Zhu et~al.(2016)Zhu, Amsler, Fuhrer, Schaefer, Faraji, Rostami,
  Ghasemi, Sadeghi, Grauzinyte, Wolverton, and Goedecker]{OMFPs2016}
Li~Zhu, Maximilian Amsler, Tobias Fuhrer, Bastian Schaefer, Somayeh Faraji,
  Samare Rostami, S.~Alireza Ghasemi, Ali Sadeghi, Migle Grauzinyte, Chris
  Wolverton, and Stefan Goedecker.
\newblock A fingerprint based metric for measuring similarities of crystalline
  structures.
\newblock \emph{Journal of Chemical Physics}, 144\penalty0 (3), 1 2016.
\newblock \doi{10.1063/1.4940026}.

\bibitem[Zhu et~al.(2021)Zhu, Wang, Shi, Ji, and Cui]{zhu2021interpreting}
Meiqi Zhu, Xiao Wang, Chuan Shi, Houye Ji, and Peng Cui.
\newblock Interpreting and unifying graph neural networks with an optimization
  framework.
\newblock In \emph{Proceedings of the Web Conference 2021}, pp.\  1215--1226,
  2021.

\end{thebibliography}
\bibliographystyle{iclr2024_conference}

\clearpage
\appendix
\section{Appendix}

\subsection{Non-locality of MFNs and Combes-Thomas Theorem}
\label{sec:combes_thomas}
The degree of non-locality of the interaction can be precisely quantified. If ${\bf H}$ has a finite interaction range, then the Combes-Thomas theorem~\cite{Combes1973} implies that
$|(z{\bf I}-{\bf H})^{-1}_{i, j} | \leq C e^{- c \gamma_z d_{ij}}$, where $\gamma_z = {\rm dist}(z, \sigma({\bf H}))$, $d_{ij}$ is the length of the shortest path from node $i$ to node $j$, and $C, c$ may depend on the norm of ${\bf H}$ but can be chosen uniformly provided $\|{\bf H}\|_\infty$ (operator max-norm) has an upper bound. Our normalisation of weights in Appendix~\ref{sec:matrix_norm} is an even stronger requirement and hence guarantees this.  \

Since we have taken ${\bf H}$ self-adjoint with real spectrum, an estimate for $\gamma_z$ is $\text{Imag}(z)$. As a result, if we constrain the poles in the parameterization of $f_\theta$ to be at $\text{Imag}(z_s) = \gamma$, then the resulting matrix function will satisfy 
\begin{equation} \label{eq:asympt_decay}
    \big| [ f_\theta({\bf H}) ]_{i,j} \big| \leq C e^{-c\gamma d_{ij}}.
\end{equation}
Therefore, the degree of locality can be controlled by constraining $\text{Imag}(z_s)$ to be some fixed value. 
While \eqref{eq:asympt_decay} only gives an upper bound, it is not difficult to see that it can in fact be attained for a suitable choice of matrix function. 
In practice, however, $f_\theta$ can be seen as approximating an unknown target $\tilde{f}$ with learnable weights $w_s$. If $\tilde{f}$ is analytic in a neighborhood of $\sigma({\bf H})$, then $ \big|[ \tilde{f}({\bf H})]_{i,j}| \leq C e^{-c\tilde{\gamma} d_{ij}},$ where $\tilde{\gamma}$ measures how smooth $\tilde{f}$ is (distance of poles of $\tilde{f}$ to $\sigma{(\bf H}$)). For our parameterization $f_\theta$ we therefore expect the same behavior $\gamma = \tilde{\gamma}$ in the pre-asymptotic regime of moderate $d_{ij}$ but transition to the asymptotic rate \eqref{eq:asympt_decay} for large $d_{ij}$.  An in-depth analysis of such effects go beyond the scope of this work. 

\subsection{Geometric expressiveness and relationship to infinite layers of MPNNs}
\label{sec:geom-expressiveness}
\vspace{-5pt}
The WL-test quantifies a network's ability to distinguish non-isomorphic graphs. Its extension to graphs embedded in vector spaces is known as the geometric WL-test~\citep{joshi2023expressive}. The expressivity in the following discussion adheres to these definitions.
In the context of graphs in $\R^3$, practical relevance is high. 
A clear relationship exists between equivariant MPNNs with linear updates and infinite layers and a one-layer MFN due to the matrix function's product structure.
The features constructed by a one-layer MFN with a single matrix channel and two-body matrix entries closely resemble the features of a two-body equivariant MPNN with \textbf{infinite layers} and \textbf{linear updates}. When matrix entries incorporate features beyond two-body, the one-layer MFN becomes more expressive than its two-body feature counterpart.

Here, we give an intuition of the relationship between the two approaches.
Any analytic matrix function $f$ admits a formal power series expansion, valid in its radius of analyticity,
\begin{equation}
    f(H) = \sum_{k=0}^{\infty} c_{k} H^{k}
\end{equation}
where $c_{k}$ are the complex (or real) coefficients of the expansion. Let us look at the diagonal elements of each of these powers that we will extract as the next doe features,
\begin{equation}
(H)^{2}_{ii, c00} = \sum_{j \in \mathcal{N}(i)} H_{ij,c0m}  H_{ji,cm0}, \quad (H)^{3}_{ii, c00} = \sum_{j \in \mathcal{N}(i)}  \sum_{k \in \mathcal{N}(j)} H_{ij,c0m} H_{jk,cmm'} H_{ji,cm'0} 
\end{equation}
The sum over $\mathcal{N}(i)$ is forced due to the sparisity of the constructed matrix $H$ in our method. Therefore, we observe a convolutional structure similar to message-passing methods. The power $2$, corresponds to a single convolution, the power $3$ to two convolutions, all the way to infinity. Each coefficient of the matrix function that is \textbf{ learnable} can act like weights in the linear update of the analogous layer. Because the power series expansion is infinite, one sees a clear analogy between the two.

\vspace{-8pt}
\subsection{Interpretability of the MFN operators}
\label{sec:interpretability}
\vspace{-6pt}
In the case of the Euclidean group, the matrices learned in MFNs have the same symmetries as Euclidean operators.
Euclidean operators play a crucial role in quantum mechanics. They are defined as self-adjoint operators (in the space of square-integrable functions) that are equivariant to the action of the group of rotations, translations, reflections, and permutations. When these operators are expended on the basis of the irreducible representations of the rotation group, they exhibit a block structure (see Appendix~\ref{sec:equivariant-op}) in which each entry is labeled with representations $(ss, ps, pp, dd,...)$. 
\begin{figure}[h!]
    \centering
\includegraphics[width=0.5\textwidth]{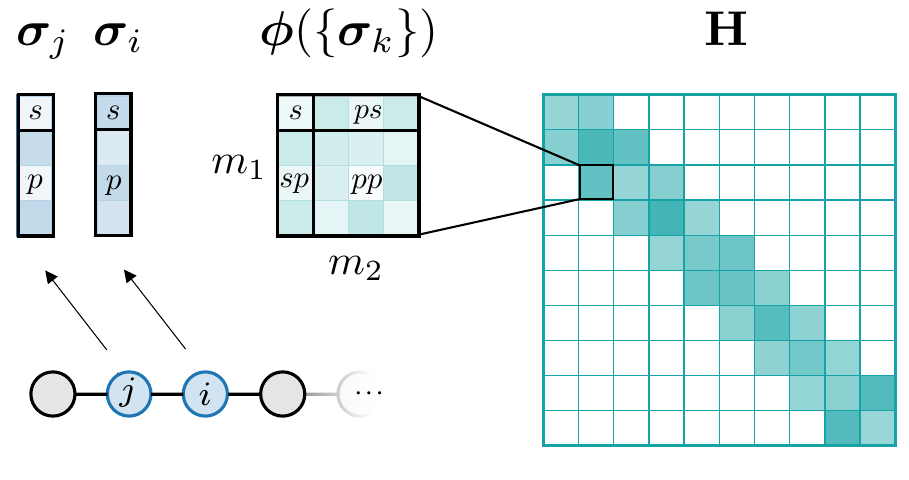}
    \caption{\textbf{Block structure} of a Euclidean Operator, H, learnt in the MFNs. Each entry in H corresponds to a different product of representations of the group of rotations  $(ss, sp, pp,...)$.}
    \label{fig:enter-label}
\end{figure}

The most central of these operators is the Hamiltonian operator. The energy of a quantum system is related to the trace of a specific matrix function of the Hamiltonian,
\begin{equation}
    E = \text{Tr}f(\bf{H})
\end{equation}
The Hamiltonian is usually computed as a fixed point of a self-consistent loop that minimizes the energy. This loop introduces many-body non-local effects.
This motivates us to use many-body functions to parameterize our matrices and to learn the fixed point directly via learning a matrix function of many matrices. This is in opposition to tight-binding methods that usually construct a single low-body Hamiltonian with physics-inspired functional forms but require several self-consistent iterations to converge. 

\subsection{Matrix normalization}
\label{sec:matrix_norm}
\vspace{-5pt}
Batch normalization~\citep{ioffe2015batch} and layer normalization~\citep{ba2016layer} are two techniques widely used in deep learning to stabilize training, improve convergence, and provide some form of regularization.

In the context of MFNs, a similar normalization strategy is needed to ensure stable training and improved convergence. Instead of normalizing the features directly as conventional normalization layers do, we normalize the eigenvalues of a set of H matrices with batch dimension $[$ batch, channels, $n$, $n$$]$, where $n$ is the size of the matrix. The aim is to adjust their mean and variance to 0 and 1, respectively, much like standardization in traditional batch or layer normalization. For the batch matrix norm, the mean is taken across the batch dimension, and for the layer matrix norm, the normalization is taken across the channel dimension.
The mean and the variance of the eigenvalues are computed using the following formulas,
\begin{equation}
    \mathbb{E}(\Lambda) = \frac{\text{tr}({\bf H})}{n}, \quad \text{Var}(\Lambda) = \frac{\text{tr}({\bf H}^2)}{n-1} - \frac{\text{tr}({\bf H})^2}{n(n-1)}
\end{equation}
This normalization of eigenvalues ensures that the spectral properties of the graph used for representation in the MFN are effectively standardized, contributing to better training stability and convergence.
\subsection{Alternatives updates for MFN}

Many updates can be defined in order to extract non-local features from the matrix function. 

\label{sec:updates-appendix}
\begin{enumerate}[leftmargin=1cm]
 \item The \textbf{dense update} utilizes the entire matrix, including all \textbf{off-diagonal terms}, to update the next matrix function, 
    \begin{equation}
    \label{eq:update-full}
        h^{(t+1)}_{icm} = V^{(t)}_{icm} + \sum_{\tilde{c}} w_{c\tilde{c}}^{(t)} f_{\theta}^{(t)}({\bf H}^{(t)}_{\tilde{c}} + f_{\theta}^{(t-1)}({\bf H}^{(t-1)}_{\tilde{c}}))_{ii, m0}
    \end{equation}
    This update can not be performed using selected inversion techniques, but it can offer additional expressiveness guarantees.
    \item The \textbf{sparse update:} uses only the parts of the matrix corresponding to the connected nodes to update the nodes and edges of the graph to obtain the matrix function in the next layer, 
    \begin{equation}
    \label{eq:update-sparse}
        h^{(t+1)}_{icm} = V^{(t)}_{icm} + \sum_{\tilde{c}} w_{c\tilde{c}}^{(t)} f_{\theta}^{(t)}({\bf H}^{(t)}_{\tilde{c}} + \{ f_{\theta}^{(t-1)}({\bf H}^{(t-1)}_{\tilde{c}})_{ij} \}_{j \in \mathcal{N}(i)})_{ii, m0}.
    \end{equation}
\end{enumerate}
\subsection{Equivariance}
\label{sec:equivariant-op}

\paragraph{Proof of equivariance of continuous matrix functions}
We will first prove the invariance of the spectrum of an operator. Let $H \in \C^{n \times n}$ be a self-adjoint matrix.  $H$ admits an eigendecomposition, $H =U\Lambda U^{T}$. The eigenvalues $\Lambda$ are the roots of the characteristic polynomial of $H$,
\begin{equation}
    \det(H - \lambda I_{n}) = 0, \quad \forall \lambda \in \Lambda
\end{equation}
Let $G$ be a reductive Lie group. Then any finite-dimensional representation of $G$, can be viewed as a representation of the maximal compact subgroup of the complexification of $G$, by the Weyl unitary trick. Therefore, they can analytically continued to unitary representations.  Let $\rho : G \to \C^{n \times n}$ be a representation of $G$ taken unitary, $\rho(g) \rho(g)^{*} = I_{n}, \forall g \in G$. Let $H_{1}$ and $H_{2}$ be two matrices related by an action of $G$, then $H_{1} = \rho(g) H_{2} \rho(g)^{*}$ for some $g \in G$. Let $\Lambda_{1}$ and $\Lambda_{2}$ be their respective spectrums. For all $\lambda \in \Lambda_{1}$
\begin{align}
\det(H_{1} - \lambda I_{n}) &= \det(\rho(g) H_{2} \rho(g)^{*} - \lambda I_{n})
=  \det(\rho(g) (H_{2}  - \lambda I_{n}) \rho(g)^{*}) \\ &= det(\rho(g) \rho(g)^{*}) det(H_{2}  - \lambda I_{n}) = det(H_{2}  - \lambda I_{n}) = 0
\end{align}
Therefore the eigenvalues of $H_{1}$ are also eigenvalues of $H_{2}$ and $\Lambda_{1} = \Lambda_{2}$.

We will now prove that if $u \in U_{1}$ is an eigenvector of $H_{1}$ with eigenvalue $\lambda$ then $\rho(g)u$ is an eigenvector of $H_{2}$. For all $u \in U_{1}$
\begin{align}
    H_{2} (\rho(g) u) &=  \rho(g) H_{1} \rho(g)^{*} (\rho(g) u) = \rho(g) H_{1} u \\
    & = \lambda (\rho(g)u)
\end{align}

Now let's prove the equivariance of matrix function. Let $f : \C \to \C$ be a continuous function, then a matrix function is defined as,
\begin{equation}
    f(H) = U f(\Lambda) U^{T}
\end{equation}
Therefore for any $g \in G$,
\begin{align}
    f(H \circ g) &= f(\rho(g) H \rho(g)^{*}) = \rho(g)U f(\Lambda)U^{T}\rho(g)^{*} \\
    &=  \rho(g)f(H)\rho(g)^{*}
\end{align}

$\hfill \square$
\paragraph{Equivariance of the resolvent}
The pole expansion yields a straightforward proof  of $f_\theta({\bf H})$ equivariance:  
\begin{equation}
\begin{split}
    f_\theta({\bf H} \circ g) 
    = \sum_s w_s \big( z_s {\bf I} - {\bm \rho} {\bf H} {\bm \rho}^* \big)^{-1} 
    = \sum_s w_s {\bm \rho} \big( z_s {\bf I} -  {\bf H}  \big)^{-1} {\bm \rho}^* 
    = {\bm \rho} f_\theta({\bf H}) {\bm \rho}^*.
\end{split}    
\end{equation}
For a general continuous function $\tilde{f}$, the analogous result follows by density.

\subsection{Multivariate Matrix function}
\label{sec:multivariate}
As MFNs extract features from multiple operators, it is useful to extend the univariate matrix function in Equation~\ref{eq:mfn-function} to a multivariate matrix function. The space of multivariate functions $\mathcal{F}(\mathcal{H}(\mathcal{G}) \times ... \times \mathcal{H}(\mathcal{G}))$ is isomorphic to the closure of the span of the tensor product of functions of single operators $\mathcal{F}(\mathcal{H}(\mathcal{G}))^{\otimes n}$,
%
%
We call the number of variables $n$, the correlation order of the matrix function. The resolvent expansion can be generalized to the multivariate matrix function, using matrix products of resolvents,
\begin{equation}
    f({\bf H}_{1}, ..., {\bf H}_{n})={\frac {1}{\left(2\pi i\right)^{n}}}\int \cdots \iint _{\partial D_{1}\times \cdots \times \partial D_{n}}{\frac {f(z_{1},\ldots ,z_{n})}{(z_{1}{\bf I}- {\bf H}_{1})\cdots (z_{n}{\bf I} - {\bf H}_{n})}}\,dz_{1}\cdots dz_{n}
\end{equation}
Multivariate matrix functions create higher order features. A one-layer MFN with a matrix function of correlation order $n$ is as expressive as a \textbf{$(n + 1)$-body} equivariant MPNN with \textbf{infinite layers} and \textbf{linear updates}.
The space $\mathcal{F}(\mathcal{H}(\mathcal{G}))^{\otimes n}$ is of very high dimension and it is possible to use standard compression techniques to find more efficient approximations such as tensor decomposition~\citep{TraceDarby2023}.
The use of linear combinations of matrices in Equation~\ref{eq:matrix-construction} approximates some of this space.
We leave further exploration of these kinds of network for future work.

\subsection{Runtime comparison}

The current MFN implementation is a prototype implementation and has a cubic scaling. However, fast inversion methods are well established, and we explain in Section~\ref{sec:resolvent} how this scaling can be reduced significantly to less than quadratic scaling, all the way to linear scaling, by exploiting the sparsity of the matrices we construct and selected inversion methods. 

Although these methods are known, they require significant effort to implement and integrate them into machine learning libraries like Pytorch due to the need for specialized CUDA kernels. We intend to make this effort and release open-source code in future work, but we believe that this is beyond the scope of this paper, where we focus on the novelty of the architecture and its expressivity. 

For the sake of completeness, we show run-time comparison on the cumulene examples between MACE and the MFN. We time the models energy and forces evaluation on an A100 GPU. The graph layer construction of the MFN model is a MACE layer with the same hyperparameters as the MACE model. We use a one layer MFN model with 16 poles and 16 matrix channels. The MACE model has 
$128$ channels, maximal angular resolution of $l_{max}=3$ and message passing equivariance $L=1$.
\begin{table}[h!t]
\centering
\caption{\textbf{Runtime comparison} of a energy and force call of a MACE model and MFN model on a 20 atoms cumulene. The MFN models constructs equivariant matrices up to $L=1$.}
\label{tab:run_time}
\vspace{-10pt}
\resizebox{0.4\linewidth}{!}{%
    \begin{tabular}{l|cccccccc}
        \toprule
        Method & MACE & MFN$^{\text{MACE}}$ \\
        \midrule
        Time (ms/atom) & 0.7 & 2.8 \\
        \bottomrule
    \end{tabular}
}
\end{table}

\subsection{Details of numerical experiments}
\label{sec:appendix-numerical}

\subsubsection{Cumulenes}



\paragraph{Dataset}

The cumulene dataset is designed to test the expressivity of graph neural networks and their non-local capabilities. The task is to regress energy and forces from 3D molecular configurations. The ground truth energy is obtained using the ORCA quantum chemistry code at the density functional theory level of accuracy using the \textit{wB97X-D3} functional, the \textit{def2-TZVP} basis set, and very tight SCF convergence criteria.

\paragraph{Length and angle scans}
\begin{figure}[h!t]
    \centering
    \includegraphics[width=0.88\linewidth]{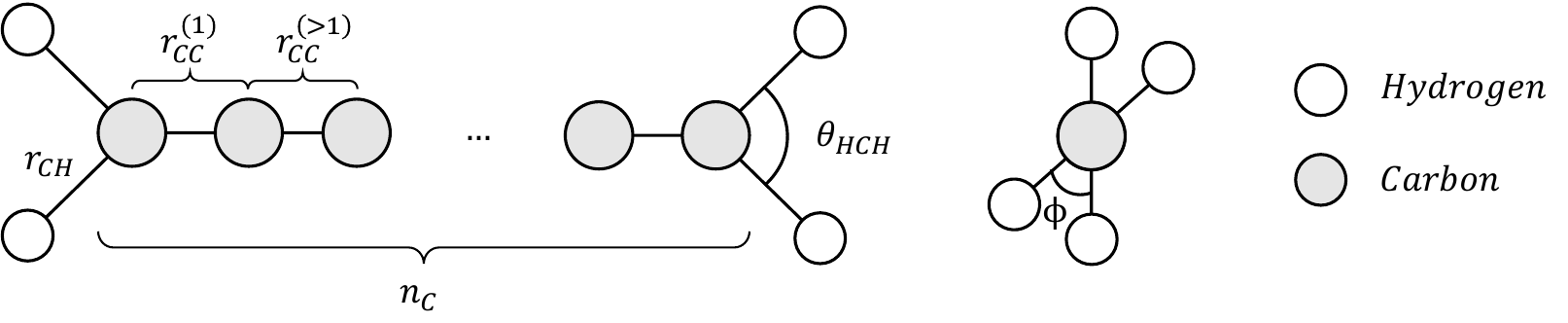}
    \caption{\textbf{Angles and distances that define a cumulene graph} used to test expressivity in Figure~\ref{fig:cumulenes-combined}. The carbon-hydrogen ($r_{CH}$), first carbon-carbon ($r_{CC}^{(1)}$), and remaining carbon-carbon distances ($r_{CC}^{(>1)}$) are set to 1.086~\AA{}, 1.315~\AA{} and 1.279~\AA{} respectively. The angle between the hydrogen-carbon-hydrogen ($\theta_{HCH}$) is fixed at 118.71 degrees and the dihedral angle $\phi$ depends on the experiment as detailed in the main text.}
    \label{fig:appendix-cumulene-graph}
\end{figure}
The configurations for the length and angle scans consist of chain-like graphs as in Figure~\ref{fig:appendix-cumulene-graph}. The length scans range from 3 to 30 atoms fixed at an angle of 5 degrees. Additionally, we scan the angle between the two terminating hydrogen groups for a cumulene with twelve carbon atoms ($n_C = 12$). At increments of 3 degrees, we scan from 0 to 90 degrees. By symmetry, this covers the entire range of unique configurations from 0 to 360 degrees. For the expressivity scan, all internal coordinates (see Figure~\ref{fig:appendix-cumulene-graph}) that uniquely define the cumulene are kept constant and set to the geometry-optimised configuration of the length 30 cumulenes. The carbon-hydrogen, first carbon-carbon, and remaining carbon-carbon distances are set to 1.086~\AA{}, 1.315~\AA{} and 1.279~\AA{} respectively (see Figure~\ref{fig:appendix-cumulene-graph}). For the length 12 cumulene, the distance between the most distant carbons is 14.1\AA{}. Thus, it becomes impossible for a purely local equivariant MPNN, with cutoff 3\AA{} and two message-passing layers, to differentiate between two graphs at different angles. 

Note that the relative energies of~\ref{fig:cumulenes-combined} are obtained by subtracting individual atom energies. These are -16.3eV and -1036.1eV for each hydrogen and carbon atoms respectively.

\paragraph{Guaranteed Non-local dataset}

The GNL dataset tests how well a model can extrapolate to an unseen number of atoms. Furthermore, configurations are rattled to see how well the model can capture both local and non-local effects. Configurations are generated starting from the relaxed geometries of cumulenes with length 0-20. Relaxations are carried out with the smaller \textit{6-31G} basis set. The cumulenes are subsequently rotated by an equally spaced scan with increment 6 degrees, starting from a random angle, and the positions are subsequently randomly perturbed by Gaussian noise with standard deviation 0.01~\AA{}. The training and validation set contain cumulenes of length 0-10 and 13-14, with 10 and 3 samples, respectively. The test set contains in-domain configurations with two samples for each of the lengths present in the training set. Furthermore, it contains configurations with 11-12 and 15-16 carbon atoms, labeled out-of-domain. In total, the train, validation, and test set contain 200, 50 and 170 configurations.

\paragraph{MACE and MFN Model}
Both the local MACE and MFN are trained on the same graphs, which means that the cutoff distance is fixed at 3\AA{}, including information from the nearest neighbor. For the MFN model, we use the architecture described in Section~\ref{sec:example-mfn}. Both models are trained with two layers, allowing the local model to distinguish changes that are separated by 12\AA{}. The MFN is trained using 16 matrix channels and 16 poles. For the matrix construction step, we use a MACE~\cite{Batatia2022mace} layer with a correlation order 3, $l_{\text{max}} = 3$, $L=1$ and $128$ channels to generate node features $h_{i}$. 
We use the diagonal update~\ref{eq:update} to update the node features of MACE and reiterate. The readout at the first layer is a linear and at the second it is a one layer MLP with 16 hidden dimensions.

\paragraph{Spookynet Model}

The Spookynet architecture was trained on the same dataset as the MFN model. We use a one-layer model with 5.5 \r{A} cutoff to reproduce the receptive field of MACE. We use the global attention and electrostatic interactions. The model has 128 channels.

\paragraph{Training}
All models underwent initial training on the GNL dataset before transfer learning was applied to the relevant length and angle scan data sets. The saved model corresponds to the epoch that exhibits the minimum loss values. Details on settings such as learning rate and epoch count are disclosed in Table~\ref{tab:model-parameters}. Training incorporated both energy and forces, with adjustable weights for each observable. In particular, for length and angle scans, an additional 100 epochs with zero force weight were used after initial training, ensuring the depiction of the minimal possible energy error in Figure~\ref{fig:cumulenes-combined}, as the lowest-loss model is saved.

\begin{table}[h]
    \centering
    \caption{Model training parameters. For the matrix functions the number of poles ($n_p$) and matrix channels ($c$) are indicated. }
    \label{tab:model-parameters}
    \resizebox{0.9\textwidth}{!}{%
    \begin{tabular}{llrlrrrll}
        \toprule
        dataset & model & epochs & lr & $E_{\text{weight}}$ & $F_{\text{weight}}$ & $n_{\text{layers}}$ & $r_{\text{max}}$ & other \\
        \midrule
        \textbf{Length Scan} & MFN (L=0) & 1240 & 1e-2 & 1000 & 100 & 2 & 3 & $n_p$=16,$c$=16 \\
        & MFN (L=1) & 5628 & 1e-5 & 1000 & 100 & 2 & 3 & $n_p$=16,$c$=16 \\
        & MACE & 5288 & 0.005 & 1000 & 100 & 2 & 3 & - \\
        \midrule
        \textbf{Angle Scan} & MFN (L=0) & 1240 & 1e-2 & 1000 & 100 & 2 & 3 & $n_p$=16,$c$=16 \\
        & MFN (L=1) & 656 & 1e-5 & 1000 & 100 & 2 & 3 & $n_p$=16,$c$=16 \\
        & MACE & 954 & 0.005 & 1000 & 100 & 2 & 3 & - \\
        \midrule
        \textbf{GNL} & MFN (L=1) & 4958 & 1e-2 & 100 & 1 & 2 & 3 & $n_p$=16,$c$=16 \\
        & MACE & 1260 & 1e-2 & 100 & 1 & 2 & 3 & - \\
        & Spookynet & 3500 & 1e-4 & 0.10 & 0.90 & 1 & 5.5 & attention \\
        \bottomrule
    \end{tabular}
    }
\end{table}

\subsubsection{ZINC}

\paragraph{Dataset}
The ZINC dataset ~\citep{Irwin2004} contains 12,000 small 2D molecular graphs with an average of 23 nodes with information on the node attributes and the edge attributes. The task is to regress the constrained solubility $\log \text{P - SA - }  \text{\#cycle}$ where $\log \text{P}$ is the octanol-water partition coefficients,
SA is the synthetic accessibility score, and $\text{\#cycle}$ is the number of long cycles. For
each molecular graph, the node features are the species of heavy elements.

\paragraph{Model}
The model is made up of six layers.
Only the first two layers are MFN layers (in order to save on the number of parameters).
The initial edge features $e_{ij}^{(0)}$ are obtained by concatenating PDF~\cite{yang2023better} descriptors used for ZINC and the input edge features of the data set.  At each layer $t$, the initial nodes features, $h_i^{(t)}$, are computed using a convolutional GNN with 140 channels. 
For the first two layers, the matrix is then formed using a multilayer perceptron (MLP) of size $[560,16]$ with the GELU activation function and a batch norm,
\begin{equation}
    H_{ij,c} = \text{MLP}(h_{i}^{(t)}, h_{j}^{(t)}, e_{ij}^{(t)})
\end{equation}
We compute the matrix function, using 16 poles and 16 channels.
We extract the diagonal elements to update the node features, and use the off diagonal elements to update the edge features by passing them into a residual 2 layers MLP of size $[512, 140, 140]$. We used an average pooling followed by a linear layer for the readout.

\paragraph{Training}
 Models were trained with AdamW,
with default parameters of $\beta_{1}$ = 0.9, $\beta_{2}$ = 0.999, and $\epsilon = 10^{-8}$ and a weight decay of $5e-5$.
We used a learning rate of 0.001 and a batch size of 64.
The learning rate was reduced using an on-plateau scheduler.

\paragraph{Baselines} 
\label{sec:baseline-zinc}
The baseline models used for the ZINC dataset comparisons include: 
GCN~\citep{kipf2017semi},
GAT~\citep{velickovic2018graph},
MPNN~\citep{gilmer2017neural},
GT~\citep{dwivedi2020benchmarking},
SAN~\citep{kreuzer2021rethinking},
Graphormer~\citep{ying2021transformers},
PDF~\citep{yang2023better}

\subsubsection{TUDatasets}

\paragraph{Dataset} We train on a subset of the TUDatasets including the MUTAG, ENZYMES, PTC-MR, PROTEINS and IMDB-B subsets. The MUTAG, ENZYMES, PTC-MR, and PROTEINS datasets are molecular datasets that contain node-level information on the molecules. The IMDB-B a social network datasets.

\paragraph{Model}
The number of layers for each subset is in Table~\ref{tab:model-parameters-pure-graphs}.
We use MFN layers only for the first 3 layers (in order to save on the number of parameters). The number of hidden channels, $n_{\text{features}}$, for each model is given in~\ref{tab:model-parameters-pure-graphs}.
The initial edge features $e_{ij}^{(0)}$ are obtained by concatenating PDF~\cite{yang2023better} descriptors used for each subset and the input edge features of the data set.  In each layer $t$, the initial node features, $h_i^{(t)}$, are computed using a convolutional GNN. 
For the first two layers, the matrix is then formed using a multilayer perceptron (MLP) of size $[4 \times n_{\text{features}},16]$ with the activation function GELU and a batch norm,
\begin{equation}
    H_{cij} = \text{MLP}(h_{i}^{(t)}, h_{j}^{(t)}, e_{ij}^{(t)})
\end{equation}
We compute the matrix function, using 16 poles and 16 channels. We extract the diagonal elements to update the node features, and use the off diagonal elements to update the edge features by passing them into a residual 2 layers MLP of size $[512, n_{\text{features}}, n_{\text{features}}]$. We used an average pooling followed by a linear layer for the readout.
\vspace{-8pt}
\paragraph{Training}
 Models were trained with AdamW,
with default parameters of $\beta_{1}$ = 0.9, $\beta_{2}$ = 0.999, and $\epsilon = 10^{-8}$ and a weight decay of $5e-5$.
We used a learning rate of 0.001 and a batch size of 64.
The learning rate was reduced using an on-plateau scheduler.

\begin{table}[h]
    \centering
    \caption{Model training parameters. For the matrix functions the number of poles ($n_p$) and matrix channels ($c$) are indicated}
    \label{tab:model-parameters-pure-graphs}
    \resizebox{0.6\textwidth}{!}{%
    \begin{tabular}{llrlll}
        \toprule
        dataset & epochs & lr & $n_{\text{layers}}$ & $n_{\text{features}}$ & other \\
        \midrule
        \textbf{MUTAG} &  301 & 1e-3  & 6 & 256 & $n_p$=16,$c$=16 \\
        \midrule
        \textbf{ENZYMES} & 201 & 1e-3 & 6 & 256 & $n_p$=16,$c$=16 \\
        \midrule
        \textbf{PTC-MR} &  151 & 1e-3 & 6 & 128 & $n_p$=16,$c$=16 \\
        \midrule
        \textbf{PROTEINS} &  451 & 1e-3 & 6 & 128 & $n_p$=16,$c$=16 \\
        \midrule
        \textbf{IMDB-B} &  301 & 1e-3 & 3 & 256 & $n_p$=16,$c$=16 \\
        \bottomrule
    \end{tabular}
    }
\end{table}

\paragraph{Baseline}
\label{sec:baseline-tu}
The baseline models used for the TU datasets comparisons include:  
	GK~\citep{shervashidze2009efficient},
	RW~\citep{vishwanathan2010graph},
	PK~\citep{neumann2016propagation},
	AWE~\citep{pmlr-v80-ivanov18a},
	PSCN~\citep{niepert2016learning},
	ECC~\citep{simonovsky2017dynamic},
	DGK~\citep{yanardag2015deep},
	CapsGNN~\citep{xinyi2018capsule},
	GIN~\citep{xu2018how},
	$k$-GNN~\citep{morris2019weisfeiler},
	IGN~\citep{maron2018invariant},
	PPGNN~\citep{maron2019provably},
	GCN$^2$~\citep{de2020natural}
	GraphSage~\citep{hamilton2017inductive},
        PDF~\citep{yang2023better}
\end{document}